\documentclass[10pt,journal,compsoc]{IEEEtran}
\ifCLASSOPTIONcompsoc
  \usepackage[nocompress]{cite}
\else
  \usepackage{cite}
\fi
\ifCLASSINFOpdf
\else
\fi

\usepackage{graphicx}
\usepackage{mathptmx} 

\usepackage{latexsym}
\usepackage{times}
\usepackage{epsfig}
\usepackage[misc]{ifsym}
\usepackage{amsmath}
\usepackage{amssymb}
\usepackage{float}
\usepackage{cuted}
\usepackage{color}

\usepackage[ruled,vlined]{algorithm2e}
\usepackage{multirow}
\usepackage{caption}
\usepackage[colorlinks,linkcolor=blue]{hyperref}
\usepackage{ragged2e} 
\newcommand{\onedot}{\ifx\@let@token.\else.\null\fi\xspace}

\newcommand{\eg}{\emph{e.g}\onedot}
\newcommand{\ie}{\emph{i.e}\onedot}

\definecolor{qmcolor}{RGB}{255,165,0}

\begin{document}
%
\title{Vision Transformer with Quadrangle Attention}
%
%
%
%

\author{
        Qiming~Zhang,~\IEEEmembership{Student Member,~IEEE},
        Jing~Zhang,~\IEEEmembership{Member,~IEEE},
        
        Yufei~Xu,~\IEEEmembership{Student Member,~IEEE},
       and~Dacheng~Tao,~\IEEEmembership{Fellow,~IEEE}
\thanks{
Q. Zhang, J. Zhang, Y. Xu, and D. Tao are with the School of Computer Science, Faculty of Engineering, The University of Sydney, Australia (email: qzha2506@uni.sydney.edu.au; jing.zhang1@sydney.edu.au; yuxu7116@uni.sydney.edu.au; dacheng.tao@gmail.com).}
}

%
%

\markboth{Journal of \LaTeX\ Class Files,~Vol.~14, No.~8, August~2015}%
{Shell \MakeLowercase{\textit{et al.}}: Bare Demo of IEEEtran.cls for Computer Society Journals}
%



\IEEEtitleabstractindextext{%
\begin{abstract}
\justifying
Window-based attention has become a popular choice in vision transformers due to its superior performance, lower computational complexity, and less memory footprint. However, the design of hand-crafted windows, which is data-agnostic, constrains the flexibility of transformers to adapt to objects of varying sizes, shapes, and orientations. To address this issue, we propose a novel quadrangle attention (QA) method that extends the window-based attention to a general quadrangle formulation. Our method employs an end-to-end learnable quadrangle regression module that predicts a transformation matrix to transform default windows into target quadrangles for token sampling and attention calculation, enabling the network to model various targets with different shapes and orientations and capture rich context information. We integrate QA into plain and hierarchical vision transformers to create a new architecture named QFormer, which offers minor code modifications and negligible extra computational cost. Extensive experiments on public benchmarks demonstrate that QFormer outperforms existing representative vision transformers on various vision tasks, including classification, object detection, semantic segmentation, and pose estimation. The code will be made publicly available at \href{https://github.com/ViTAE-Transformer/QFormer}{QFormer}.

\end{abstract}

}

\maketitle

\IEEEdisplaynontitleabstractindextext

%
\IEEEpeerreviewmaketitle

\IEEEraisesectionheading{\section{Introduction}\label{sec:introduction}}

\IEEEPARstart{V}{ision} transformers (ViT) \cite{vit} have emerged as a promising approach for various vision tasks. ViT treats 2D images as 1D sequences by dividing the input image into patches and embedding them as tokens, which are then processed with stacked transformer blocks containing self-attention and feed-forward networks. Despite its simplicity, this architecture yields superior performance. However, the quadratic complexity of the vanilla self-attention over the input token length poses a challenge for processing high-resolution images. To address this issue, local window-based attention~\cite{liu2021swin} is proposed, which partitions images into several non-overlapping squares (\ie, windows) and performs attention within each window individually. This design balances performance, computation complexity, and memory footprint, thereby significantly expanding the usage of plain and hierarchical transformers \cite{liu2021swin,li2022exploring} in various vision tasks~\cite{liu2021swin,yang2021focal,dong2021cswin,zhang2022vitaev2,yang2022modeling,xu2021vitae,xu2022vitpose,wang2021kvt}. However, it also imposes a constraint on the design form of windows, \ie, the squares, thereby limiting the transformer's ability to model long-range dependency as well as handle objects of varying sizes, shapes, and orientations, which is crucial for vision tasks.

Previous studies have focused on improving window-based attention by enabling it to model long-range dependencies with advanced designs. One simple approach, as explored in the Swin transformer~\cite{liu2021swin}, involves enlarging window sizes from 7 $\times$ 7 to 32 $\times$ 32 \cite{swinv2} to include more tokens in the calculation, albeit at a higher computational cost. Other works try to manually design various forms of windows, such as Focal attention~\cite{yang2021focal}, which incorporates coarse granularity tokens to capture long-range context, cross-shaped window attention~\cite{dong2021cswin}, which uses two cross rectangular windows to model long-range dependency from both vertical and horizontal directions, and Pale~\cite{wu2021pale}, which attends to tokens in dilated vertical/horizontal directions to model long-range dependency from diagonal directions. These methods have improved image classification performance by enlarging the attention distance. However, all of them adopt fixed rectangular windows for attention calculation, despite the arbitrary sizes, shapes, and orientations of targets in images. This data-agnostic and hand-crafted window design may be suboptimal for vision transformers. In this study, we propose a data-driven solution to this problem by extending the default windows from rectangular shapes to quadrangles, the optimal parameters of which (\eg, shape, size, and orientation) can be automatically learned. It enables transformers to learn better feature representation to handle diverse objects, \eg, by capturing rich context from long-range tokens enclosed by the adaptive quadrangles.

\begin{figure*}[!ht]
    \centering
    \includegraphics[width=1\linewidth]{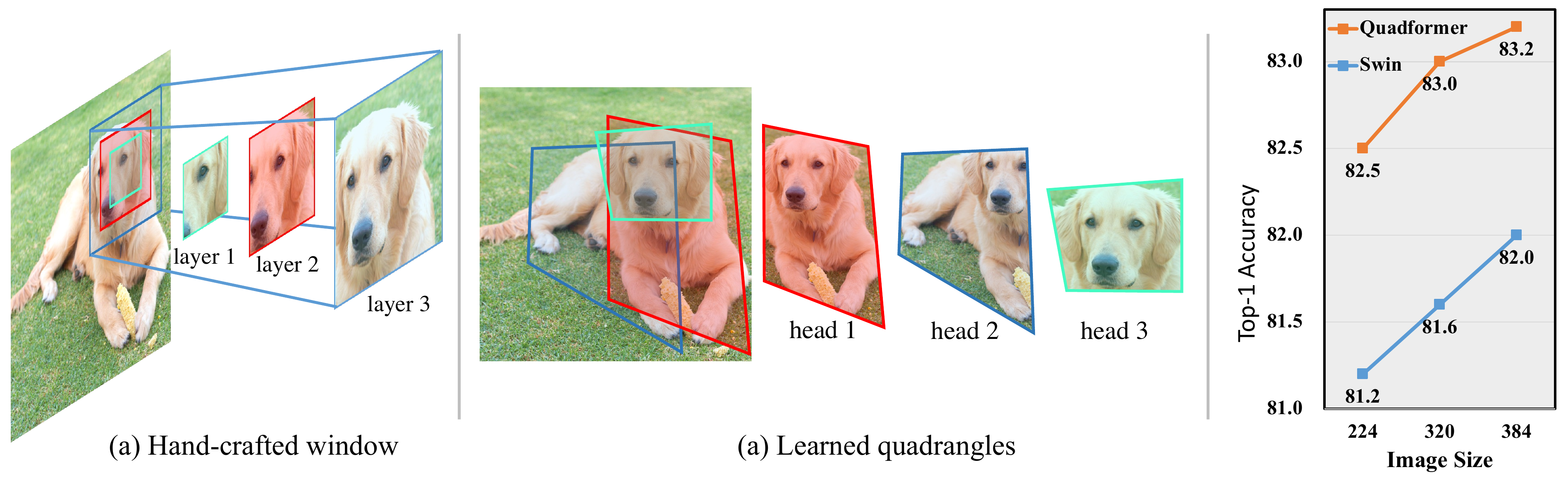}
    \caption{Comparison of window configurations in previous hand-crafted designs~\cite{liu2021swin} (a) and the proposed learnable quadrangle design (b), as well as their image classification performance under different settings of input size on the ImageNet~\cite{deng2009imagenet} validation set (c).}
    \label{fig:opening}
\end{figure*}

Specifically, we present a novel attention method dubbed \textbf{Q}uadrangle \textbf{A}ttention (QA), which aims to learn adaptive quadrangle configurations from data for calculating local attention. It employs default windows to partition the input images and uses an end-to-end learnable quadrangle regression module to predict a parameterized transformation matrix for each window. The transformation involves translation, scaling, rotation, shear, and projection, which is utilized to transform the default window to a target quadrangle. To enhance the training stability and allow good explainability, the transformation matrix is formulated as the composition of several basic transformations. Unlike window attention, where the window definition is shared between different heads in the multi-head self-attention (MHSA) layer, the proposed quadrangle transformation is performed independently for each head. This design enables attention layers to model diverse long-term dependencies and facilitates information exchange among overlapped windows without requiring window shift~\cite{liu2021swin} or token permutation~\cite{huang2021shuffle}.
We integrate QA into plain and hierarchical vision transformers to create a new architecture named QFormer, which offers minor code modifications and negligible extra computational cost. We have conducted extensive experiments and ablation studies on public benchmarks for various vision tasks including
classification, object detection, semantic segmentation, and pose estimation. The results demonstrate the effectiveness of QA and the superiority of QFormer over existing representative vision transformers. As shown in Figure~\ref{fig:opening}, the proposed QFormer significantly outperforms the Swin Transformer~\cite{liu2021swin} for image classification under different settings of input size.

In summary, the contribution of this study is threefold. (1) We present a novel quadrangle attention method that can directly learn adaptive quadrangle configurations from data. It breaks the constraint of the fixed-size window in existing architectures and makes it possible for transformers to adapt to objects of various sizes, shapes, and orientations in a much easier way. (2) We employ the QA in both plain and hierarchical vision transformers to create a new
architecture named QFormer, which offers minor code modifications and negligible extra computational cost. (3) Extensive experiments on public benchmarks demonstrate the effectiveness of QA and the superiority of QFormer over representative vision transformers on various visual tasks, including image classification, object detection, and semantic segmentation.

\section{Related Work}
\subsection{Vision transformer}
Vision transformers~\cite{vit} have demonstrated superior performance in many vision tasks by modeling long-term dependency among image patches (a.k.a. tokens)~\cite{xu2022vitpose,jing2020dynamic}. However, the use of vanilla full attention in vision transformers can lead to inferior training efficiency due to the lack of inductive bias. To address this issue, previous works~\cite{touvron2020training,xu2021vitae,dai2021coatnet,yan2021contnet} have introduced inductive bias into vision transformers either implicitly or explicitly and have achieved superior classification performance. Multi-stage designs have also been explored to better adapt vision transformers to downstream vision tasks \cite{wang2021pyramid,wang2021pvtv2,liu2021swin,wang2021crossformer,zhang2022vitaev2}. Among them, Swin Transformer~\cite{liu2021swin} is a representative work, which partitions the tokens into non-overlapping windows and conducts attention within each window to reduce the computational cost and memory footprint, especially when dealing with larger input images. However, window-based attention imposes a constraint on the attention distance due to the fixed window size. To recover the transformer's ability to model long-term dependency, various techniques have been explored, such as using additional tokens for efficient cross-window feature exchange or designing other forms of windows to allow the transformer layers to attend to far-away tokens in specific directions~\cite{fang2021msg,dong2021cswin,wu2021pale,huang2021shuffle}. These techniques rely on hand-crafted windows for attention computation and need to stack the transformers layers sequentially to enable feature exchange across all windows and model long-term dependency. As a result, they lack the flexibility to adapt well to inputs of various sizes.
In contrast, the proposed QA learns the window configurations from input features and calculates attention within such generated adaptive quadrangles. It allows transformer layers to model long-term dependency, capture rich context, and promote cross-window information exchange from the diverse ``windows'', thereby learning better feature representation for vision tasks.

\subsection{Deformable sampling}

Deformable sampling has been extensively studied to aid convolutional networks \cite{dai2017deformable,zhu2019deformable} in selectively attending to regions of interest and extracting high-quality features. Such techniques have been adapted in deformable-DETR \cite{zhu2021deformable} to help transformer detectors efficiently identify and utilize token features crucial for object detection. More recently, DPT \cite{chen2021dpt} has proposed deformable patch merging layers based on the Pyramid Vision Transformer (PVT) architecture \cite{wang2021pyramid} to enable transformers to better preserve features following downsampling. From a different perspective, our QA introduces learnable quadrangle-based window attention into transformers. By learning the position, size, orientation, and shape of the windows for attention calculation, QA circumvents the restriction of hand-crafted fixed-size windows, facilitating better adaptation of window-based transformers to objects of varying sizes, shapes, and orientations.

\begin{figure*}
    \centering
    \includegraphics[width=\linewidth]{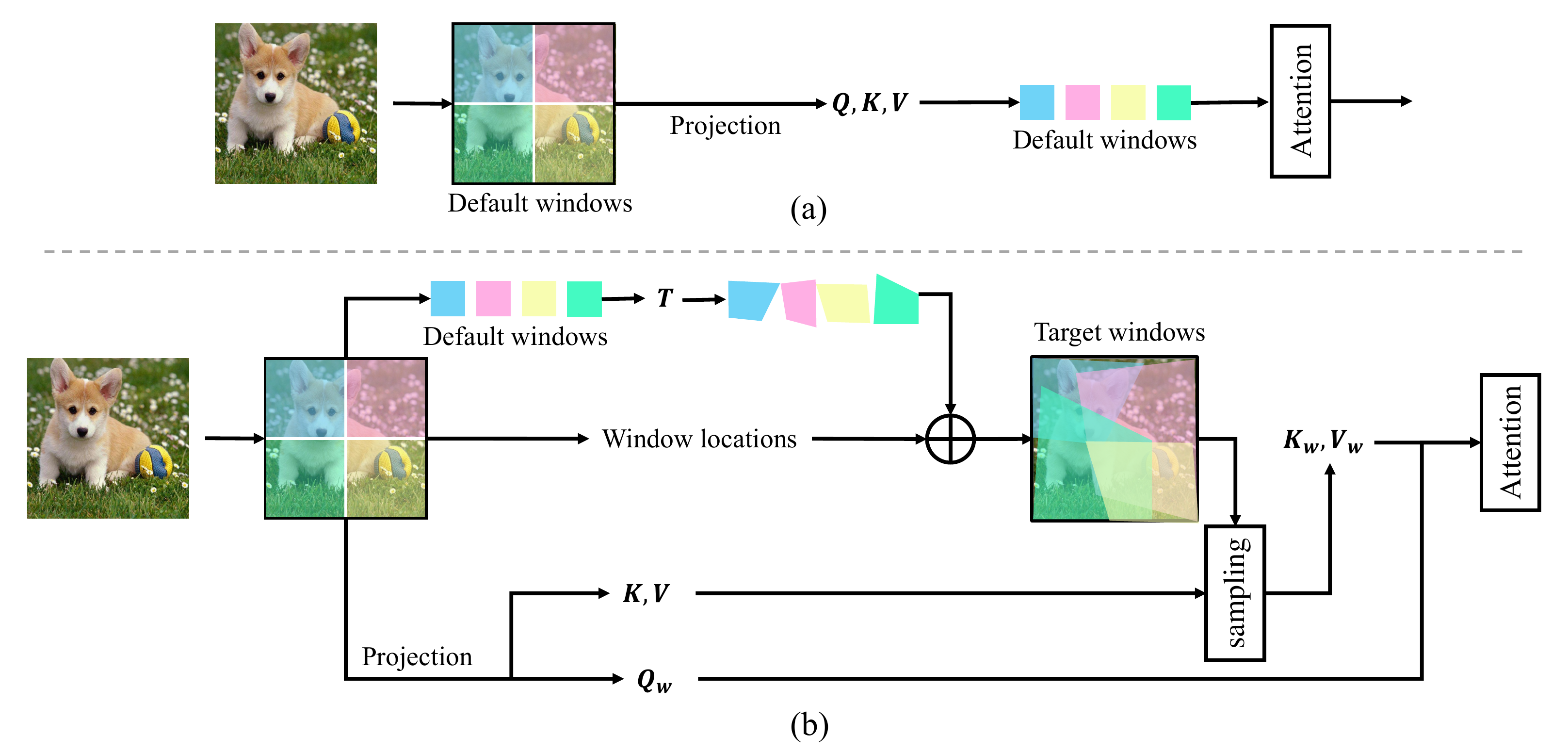}
    \caption{Illustration of window attention (a) and the proposed quadrangle attention (QA) (b).}
    \label{fig:pipeline}
\end{figure*}

\subsection{Comparison to the conference version}

A preliminary version of this paper is presented in VSA \cite{zhang2022vsa}, which learns varied-size windows for attention calculation. This paper extends the previous study with three major improvements:
\begin{enumerate}
    \item VSA uses rectangular to represent the attention area, which may not well adapt to the complex scenes in images, such as the objects of different shapes, and restrict the modeling ability. We rethink the idea from a more general concept by extending the regular rectangles to quadrangles and propose the QA accordingly. Thanks to the more flexible configuration, QA can learn the position, size, orientation, and shape of the windows for attention calculation and thus facilitate extracting the rich context of objects and learning better feature representation, while the learned rectangles in VSA can be regarded as a specific case in QA.
    \item We resort to projective transformation to transform the default windows to arbitrary quadrangles, where the transformation matrix is directly learned from data. The transformation is formulated as the composition of several basic transformations. Besides, we propose a regularization that encourages the quadrangles to cover reasonable areas to stabilize the training process and accelerate the model convergence.
    \item We employ the QA in both plain and hierarchical vision transformers to create a new architecture named QFormer, bringing only minor code modifications and negligible extra computational cost. We have conducted extensive experiments on four vision tasks including classification, object detection, semantic segmentation, and pose estimation to evaluate QA and QFormer. The results demonstrate that QFormer outperforms representative vision transformers by a large margin while enjoying low computation complexity and memory footprint, establishing it as a promising architecture for various vision tasks.
\end{enumerate}
Besides, more experiment results, ablation studies, and analyses are presented. Some visual results are also provided to demonstrate the promising performance and explainability of QA.

\section{Method}

In this section, we present the preliminaries of vision transformers, the technical details of QA, the implementation of QFormer by employing QA in both plain ViTs and hierarchical ones such as the Swin Transformer, and the computational complexity analysis.

\subsection{Preliminary}
We will first briefly review the typical window-based attention in vision transformers~\cite{liu2021swin,li2022exploring}. As illustrated in Figure \ref{fig:pipeline}(a), given the input feature map $X \in \mathcal{R}^{H \times W \times C}$ as input, it is partitioned into non-overlapping windows, \ie, $\{X^i \in \mathcal{R}^{w \times w \times C} | i \in [1, \dots, \frac{H\times W}{w^2}]\}$, where $w$ is the predefined window size and the windows are typical squares. The partitioned tokens within each window are flattened along the spatial dimension and projected to query, key, and value tokens following the same projection process of vanilla self-attention, \ie, 
\begin{equation}
    \{Q, K, V\} = \{Q^i, K^i, V^i \in \mathcal{R}^{w^2 \times C} | i \in [1, \dots, \frac{H\times W}{w^2}]\},
\end{equation}
where $Q,K,V$ represent the query, key, and value tokens, respectively. $C$ is the channel dimension. To allow the model to capture context information from different representation subspaces, the tokens are chunked along the channel dimension equally by $N$ heads, resulting in the tokens:
\begin{equation}
    \{Q_h, K_h, V_h\} = \{Q_h^i, K_h^i, V_h^i \in \mathcal{R}^{w^2 \times C'} | i \in [1, \dots, \frac{H\times W}{w^2}]\},
\end{equation}
where $h \in [1, \dots, N]$ and $C'$ is the channel dimension along each head, \ie, $C'=C/N$. We omit the subscript $h$ in the following for simplicity. Given the flattened query $Q^i$, key $K^i$, and value $V^i$ tokens from the $i$th window, the window-based attention layers conduct the self-attention as follows, \ie,
\begin{equation}
    F^i = SA(Q^i, K^i, V^i).
\end{equation}
$F^i \in \mathcal{R}^{w^2\times N \times C'}$ is the output features after the self-attention operation $SA(\cdot)$~\cite{vit}. Then the features are concatenated along the spatial and channel dimension respectively to recover the feature map. Note that the tokens within each window are processed in the same manner and we dismiss the window index notation $i$ of query, key, and value tokens in the following for simplicity. The attention operation is computed as a weighted sum of the values, where the weight, \ie, attention matrix, is determined by the similarity between the query and corresponding keys via a dot product and softmax functions, \ie,
\begin{equation}
    F = (softmax(Q K^{T}) + r)V,
    \label{eq:attention}
\end{equation}
where $r$ is the relative position embedding \cite{liu2021swin,li2021improved} to encode spatial information. It is always learnable during training.

One key advantage of the window-based attention over the villain self-attention is reducing the computational complexity to linear regards the input size, \ie, each window attention's complexity is $\mathcal{O}(w^4C)$ and the computation complexity of window attention for each image is $\mathcal{O}(w^2HWC)$. To encourage the information exchange between different windows, a shifted window strategy is used between two adjacent transformer layers in Swin~\cite{liu2021swin} and several vanilla self-attention layers are employed at intervals in ViTDet \cite{li2022exploring}. As a result, the receptive field of the model is enlarged by stacking transformer layers in sequence. However, current window-based attention restricts the attention area of the tokens within the hand-crafted fixed-size window at each transformer layer. It limits the model's ability to capture far-away context and learn better features to represent objects of different sizes, orientations, and shapes. Thus, they need to carefully tune the size of the windows for different tasks, \eg, enlarging the window size in Swin Transformers~\cite{liu2021swin,swinv2} when the input resolution becomes larger.

\begin{figure}
    \centering
    \includegraphics[width=0.9\linewidth]{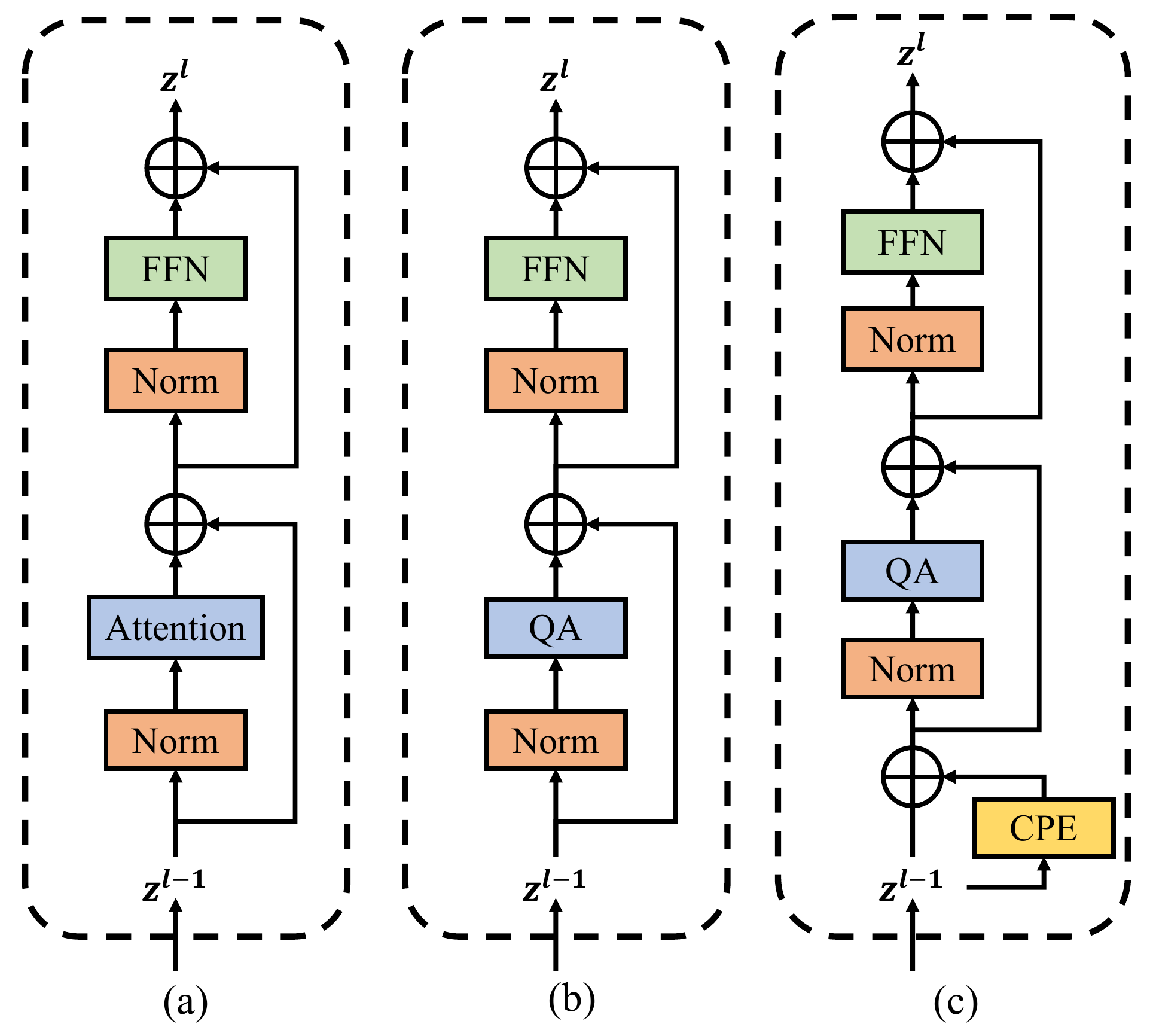}
    \caption{The block of traditional window attention (a), the proposed QA in the plain ViT (b), and the hierarchical ViT (c).}
    \label{fig:blocks}
\end{figure}

\begin{figure*}
    \centering
    \includegraphics[width=\linewidth]{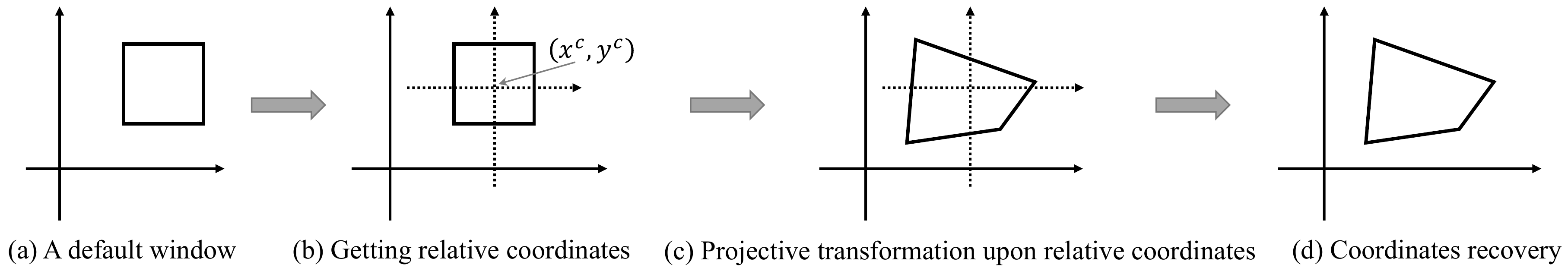}
    \caption{Illustration of the projective transformation pipeline. (a) a default window. (b) The relative coordinates with respect to the window center $(x^c,y^c)$ are calculated. (c) The target quadrangle is obtained after projective transformation upon the relative coordinates. (d) The absolute coordinates are recovered by adding the window center coordinates.}
    \label{fig:transformation}
\end{figure*}

\subsection{quadrangle attention}
\noindent\textbf{Base window generation.}
To alleviate the difficulties in using hand-crafted windows to deal with various objects, we propose the QA that allows the model to dynamically determine the appropriate location, size, orientation, and shape of each window in a data-driven manner. QA is easy-to-implement and only needs to make minor modifications to the basic structure of window attention-based vision transformers by simply substituting the attention module, as shown in Figure~\ref{fig:blocks}. 
Since the proposed QA conducts the same and independent operation for each head, we take one head as an example in the following. 
Technically, given the input feature map, it is first partitioned into several windows $X$ with the predefined window size $w$, which is the same as the window-based attention, as shown in Figure~\ref{fig:pipeline}(b). We refer to these windows as base windows and get the query, key, and value tokens from the feature from each window, respectively, \ie, 
\begin{gather}
    Q, K, V = Linear(X).
\end{gather}
We directly use the query tokens for QA calculation, \ie, $Q_w = Q$, while reshaping the key and value tokens into feature maps for the following sampling step in QA calculation. 

\noindent\textbf{Quadrangle generation.}
We regard the base windows as references and resort to the projective transformation to transform each base window into a target quadrangle. Since projective transformation does not preserve parallelism, length, and orientation, the obtained quadrangles are very flexible with respect to locations, sizes, orientations, and shapes, making them well-suited to cover objects of different sizes, orientations, and shapes. We will briefly introduce the definition of projective transformation as follows. It is represented by a transformation matrix of eight parameters:
\begin{equation}
T =
\begin{pmatrix}
    a_1 & a_2 & b_1 \\
    a_3 & a_4 & b_2 \\
    c_1 & c_2 & 1 \\
\end{pmatrix},
\end{equation}
where $a=[a_1, a_2; a_3, a_4]$ defines the transformation of scaling, rotation, and shearing, $b=[b1;b2]$ defines the translation, and $c=[c1,c2]$ is a projection vector that defines how the perceived objects change when the viewpoint of the observer changes in the depth dimension. 

As shown in Figure \ref{fig:pipeline} (b), given the tokens in a base window $X_w$, QA uses a quadrangle prediction module to predict the projective transformation regarding the base window. However, directly regressing the eight parameters of the projective matrix is not easy. Instead, we disentangle the projective transformation into several basic transformations and predict the parameters for each of the basic transformations accordingly. Specifically, the quadrangle prediction module first predicts the surrogate parameters $t \in \mathcal{R}^9$, where the module consists of an average pooling layer, a LeakyReLU~\cite{xu2015empirical} activation layer, and a $1 \times 1$ convolutional layer in sequence:
\begin{equation}
    t = Conv \circ LeakyReLU \circ AveragePool(X_w).
    \label{eq:t}
\end{equation}
Then, it obtains the basic transformations including scaling $T_s$, shearing $T_h$, rotation $T_r$, translation $T_t$, and projection $T_p$ based on the output $t$ in Eq.~\eqref{eq:t}:
\begin{equation}
T_s =
\begin{pmatrix}
    t_1+1 & 0 & 0 \\
    0 & t_2+1 & 0 \\
    0 & 0 & 1 \\
\end{pmatrix},
T_h =
\begin{pmatrix}
    1 & t_3 & 0 \\
    t_4 & 1 & 0 \\
    0 & 0 & 1 \\
\end{pmatrix},
\end{equation}

\begin{equation}
T_r =
\begin{pmatrix}
    \cos{t_5} & -\sin{t_5} & 0 \\
    \sin{t_5} & \cos{t_5} & 0 \\
    0 & 0 & 1 \\
\end{pmatrix},
\end{equation}

\begin{equation}
T_t =
\begin{pmatrix}
    1 & 0 & \beta_1 t_6 \\
    0 & 1 & \beta_2 t_7 \\
    0 & 0 & 1 \\
\end{pmatrix},
T_p =
\begin{pmatrix}
    1 & 0 & 0 \\
    0 & 1 & 0 \\
    t_8 & t_9 & 1 \\
\end{pmatrix},
\end{equation}
where $\beta_1 = \frac{W}{w}$ and $\beta_2 = \frac{H}{w}$ scale the translation regarding the image size to help adapt the model to different input sizes. 
Finally, the transformation matrix $T$ is obtained by multiplying all transformations sequentially, \ie,
\begin{equation}
    T = T_s \times T_h \times T_r \times T_t \times T_p.
\label{eq:basic composition}
\end{equation}

Given the estimated projection matrix, we get the location of the projected points via the standard projection process, \ie, given the coordinates of a point $(x_1,y_1)$, the transformation can be done by the simple multiplication:
\begin{equation}
    [x_{tmp}, y_{tmp}, z_{tmp}]^T = T \times [x_1,y_1,1]^T,
    \label{eq:quadrangle generation}
\end{equation}
and the final coordinates are: 
\begin{equation}
    (x_2, y_2) = (x_{tmp}/z_{tmp}, y_{tmp}/z_{tmp}).
    \label{eq:quadrangle generation normalization}
\end{equation}
The calculation is conducted for each point in the base window in parallel to get the target location in the projected quadrangle. 
It is noted that this formulation also includes the traditional window attention a special case of QA, where $t=0$ and thus $T$ is an identity matrix. By default, we initialize the weights of the quadrangle prediction module to produce $t=0$.

\begin{figure}
    \centering
    \includegraphics[width=1\linewidth]{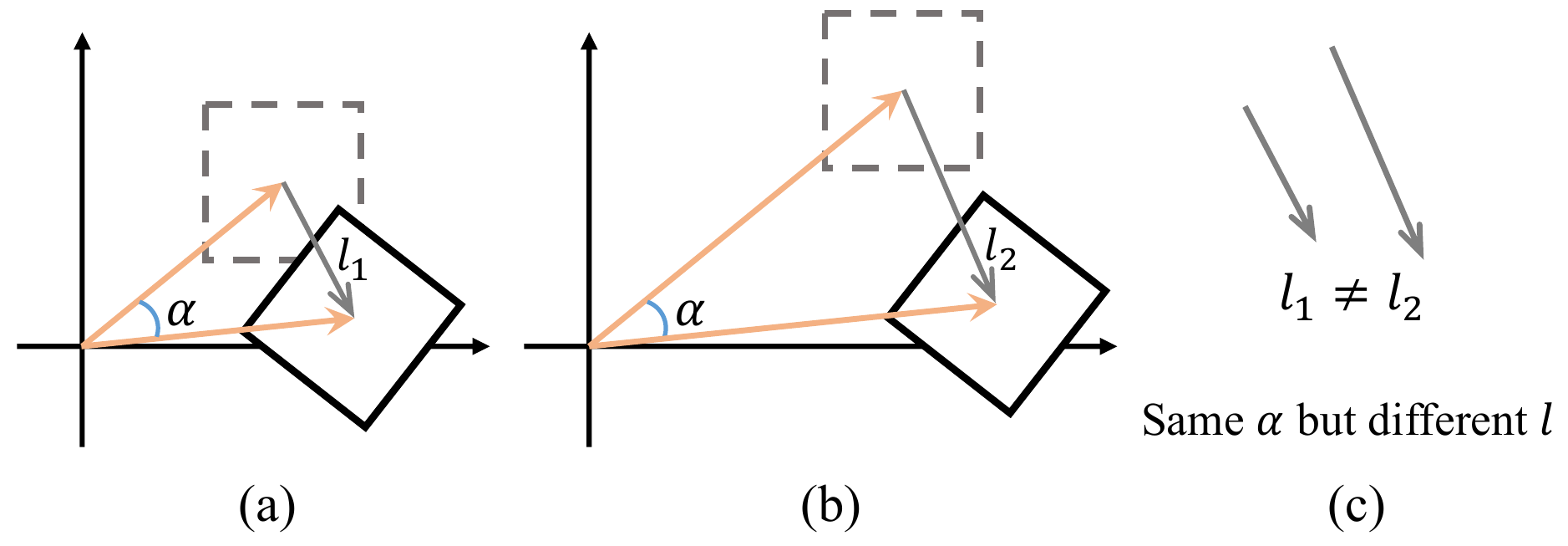}
    \caption{Comparison of the same transformation to two windows at different locations using the absolute coordinates.}
    \label{fig:ambiguity}
\end{figure}

\begin{figure*}[h]
    \centering
    \includegraphics[width=\linewidth]{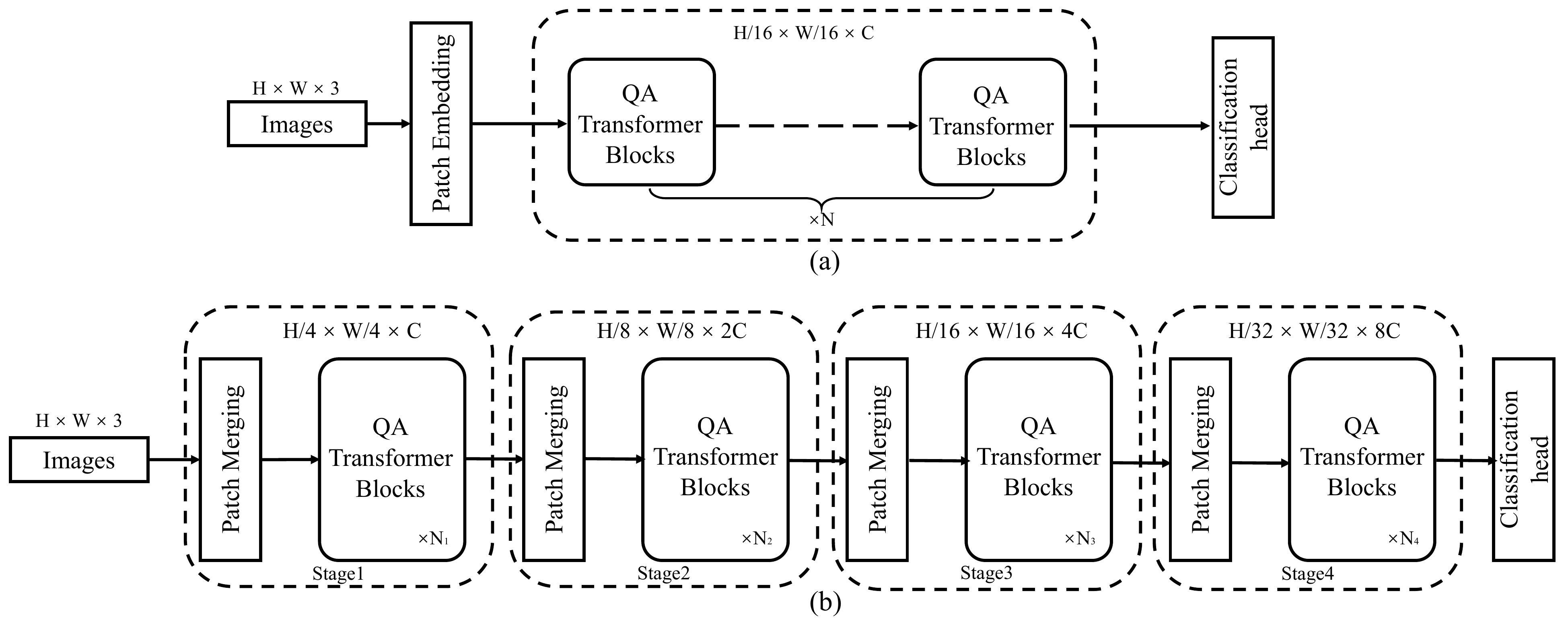}
    \caption{The pipeline of window attention (a), our proposed quadrangle attention (QA) (b), and the details of transformer blocks with the proposed quadrangle attention.}
    \label{fig:model}
\end{figure*}

However, simply using the token coordinates under the same coordinate system can lead to ambiguity when generating quadrangles as shown in Figure \ref{fig:ambiguity}. For example, given the two windows at different locations, the window far from the origin point in (b) can have a significantly different translation compared with the one close to the origin point in (a), even though they have the same projective transformation matrix, \ie, rotation. 
This will lead to optimization difficulty of the quadrangle regression module during training. To address this issue, the projective transformation is conducted by taking each window's relative coordinates as input instead of using the absolute ones as shown in Figure \ref{fig:transformation}.
Specifically, given the coordinates of tokens within a window $\{(x_i,y_i)|i=0\dots M\}$, where $i$ indexes the token and $M=w\times w$, we transform its coordinates to the relative coordinates as:
\begin{equation}
        (x^{r}_i, y^{r}_i)= (x_i - x^{c}, y_i - y^{c}),
\end{equation}
where $x^c,y^c$ denotes the center of window center as shown in Figure \ref{fig:pipeline}(b) and $(x^{r}_i,y^{r}_i)$ are the relative coordinates with respect to the center. 
After obtaining the transformation matrix, we can get the relative coordinates of each token in the target quadrangle $(x_{q,i}^r,y_{q,i}^r)$ from $(x^{r}_i, y^{r}_i)$ according to Eq.~\eqref{eq:quadrangle generation} and Eq.~\eqref{eq:quadrangle generation normalization}. 
Then the absolute coordinates after transformation are obtained as follows:
\begin{equation}
    (x_{q,i}, y_{q,i}) = (x_{q,i}^r + x^{c}, y_{q,i}^r + y^{c}).
\end{equation}

After getting the coordinates of tokens in each target quadrangle, we use the grid-sampling function to sample the key and value tokens $K_w, V_w$ from $K,V$, respectively. Thanks to the data-driven learning mechanism of window configurations in QA, there may be many diverse and overlapped quadrangles, thereby promoting the cross-window information exchange. However, such a design may generate quadrangles that cover the area outside the feature map. A simple sampling strategy is proposed to address this issue: (1) for the sampled coordinates within the feature map, a bi-linear interpolation is employed to sample the values; and (2) for those whose coordinates are outside the feature map, we use zero vector as their sampling values. Finally, we use the sampled $K_w, V_w$ and the original $Q_w$ for self-attention calculation according to Eq.~\eqref{eq:attention}.

\subsection{Regularization}
As stated above, the learned quadrangles may cover regions that are outside the feature maps and zero values are sampled in this case. Such a phenomenon hinders the learning of the quadrangle regression module since the gradients from such regions are always zero. 
To address this issue, we design a regularization term to encourage the projected quadrangles to cover more valid regions within the feature maps. Specifically, given the token coordinates $(x_{q,i}, y_{q,i})$ within a quadrangle, we define a penalty function to penalize those coordinates outside the feature map:
\begin{equation}
    R(x) = \begin{cases}
    -\lambda, & x<-1 \\
    0, & -1<=x<=1 \\
    \lambda, & x>1 \\
\end{cases}
\end{equation}
and the regularization loss for all the token coordinates is:
\begin{equation}
L_{reg} = \sum_i R(x_{q,i}) \cdot x_{q,i} + R(y_{q,i}) \cdot y_{q,i}.
\end{equation}
where $\lambda$ is the hyper-parameter. We sum the regularization loss and the common training loss, \eg, the cross-entropy loss for image classification. 

\begin{table}[htbp]
  \centering
  \caption{The specifications of QFormer. The subscript $h$ and $p$ denote hierarchical and plain architectures, respectively.}
    \begin{tabular}{cccccc}
    \hline
    Model & Layers & Channel & MLP & Heads & Params \\
    \hline
    QFormer$_h$-T &  2,2,6,2 &   96  &  4$\times$    &   3,6,12,24     & 29 M \\
    QFormer$_h$-S &  2,2,18,2 &   96  &  4$\times$  &  3,6,12,24     &  50 M \\
    QFormer$_h$-B &  2,2,18,2 &   128  &  4$\times$  &  4,8,16,32     & 88  M \\
    QFormer$_p$-B &   12    &   768    &   4$\times$    &   12  &  86 M \\
    \hline
    \end{tabular}%
  \label{tab:model specification}%
\end{table}%

\subsection{Model specifications}
We integrate QA into both plain and hierarchical vision transformers to create a new
architecture named QFormer, whose model specifications are shown in Table~\ref{tab:model specification}. Both architectures are composed of a series of transformer layers as shown in Figure~\ref{fig:model}. The hierarchical architecture QFormer$_h$ downsamples the feature maps gradually in several stages, \ie, by a factor of $4\times$, $2\times$, $2\times$, $2\times$, respectively, as shown in Figure~\ref{fig:model}(b). We use Swin Transformer \cite{liu2021swin} as the reference architecture and adopt the conditional position embedding (CPE) in CPVT~\cite{chu2021conditional} before the attention layers to encode the spatial information into the model as shown in Figure~\ref{fig:blocks} (c), \ie, 
\begin{equation}
    X = Z^{l-1} + CPE(Z^{l-1}).
\end{equation}
The number of transformer layers and the channel dimension in each stage are the same as Swin Transformer, but we remove the window shifting operation for simplicity. For large models, we use plain ViT as the reference architecture and adopt the MAE pre-trained \cite{he2021masked} weights for initialization. As shown in Figure~\ref{fig:model}(a), The plain architecture QFormer$_p$ has the same feature size for all transformer layers, following the setting in ViT \cite{vit}.

\subsection{Computation complexity analysis}
The extra computations brought by QA come from $CPE$ and the quadrangle prediction module, while the other parts, including the window-based MHSA and FFN, are exactly the same as the reference architectures. Given the input features $X \in \mathcal{R}^{H \times W \times C}$, QA in hierarchical architectures firstly uses a depth-wise convolutional layer with a $7 \times 7$ kernel to generate CPE, which brings extra $\mathcal{O}(49\cdot HWC)$ computations. For learning the projective matrix, we first employ an average pooling layer with a kernel size and a stride equal to the window size to aggregate features from the base windows, whose computational complexity is $\mathcal{O}(HWC)$. The following activation function does not introduce extra computations, and the last convolutional layer with a kernel size $1 \times 1$ takes $X_{pool} \in \mathcal{R}^{\frac{H}{w} \times \frac{W}{w} \times C}$ as the input and predicts the transformation parameters $\{t_{i} | i=1...9\}$ for each head. Thus, its computational complexity is $\mathcal{O}(\frac{HW}{w^2} \cdot 9N C)$. After obtaining the transformation matrix, we transform the default windows to quadrangles where we uniformly sample $w \times w$ tokens. Here we ignore the computational complexity of calculating the new coordinates because it is very marginal (\ie, $27HW$). The computational complexity of sampling in each quadrangle is $w^2 \times 4 \times C$, and the total computational complexity of sampling is $\mathcal{O}(4\cdot H W C)$. Thus, the total extra computations brought by QA is $\mathcal{O}\{(54 + \frac{4N}{w^2}) H W C\}$, which is far less ($\leq5\%$) than the total computations of the reference architectures. Note that the computational complexity of FFN is $\mathcal{O}(2 \alpha H W C^2)$, where $\alpha$ is the expansion ratio and $C$ is always larger than 96 for both plain and hierarchical vision transformers. When the model size increases with a larger token dimension ($C$), the extra FLOPs brought by QA can be negligible, \ie, 0.1\% for QFormer$_p$-B with $C=768$.

\section{Experiments}
\subsection{Experiment settings}
\subsubsection{Datasets}
To comprehensively evaluate the effectiveness of QA, we conduct experiments on various vision tasks, including classification, semantic segmentation, instance segmentation, detection, and pose estimation, on well-known public datasets such as ImageNet-1k~\cite{deng2009imagenet}, ADE20k~\cite{zhou2017scene}, and MS COCO~\cite{lin2014microsoft}.

\noindent\textbf{ImageNet-1k~\cite{deng2009imagenet}} is a widely used large-scale image classification dataset, organized hierarchically into 1,000 categories based on the WordNet hierarchy. It contains 1 million images for training and 50,000 for testing, with 1,000 and 50 images per category, respectively. It is a \textit{de facto} standard benchmark for evaluating the performance of backbone networks~\cite{he2016deep,liu2021swin}. Additionally, it serves as a fundamental dataset for pre-training various backbone models in either supervised or self-supervised ways. In this study, we adopt the ImageNet-1k pre-trained backbone for downstream tasks including object detection, semantic segmentation, and pose estimation, following the common practice.

\noindent\textbf{ADE20k}\cite{zhou2019semantic} is a popular dataset for semantic segmentation, comprising 27k images with 150 labeled classes. It includes 25k images for training and 2k for validation, with diverse scenarios encompassing indoor and outdoor city street scenes. In this study, we adopt ADE20k as the representative dataset for semantic segmentation, in line with previous works such as BeiT \cite{beit} and Swin Transformer~\cite{liu2021swin}.

\noindent\textbf{MS COCO}~\cite{lin2014microsoft} is a popular dataset used for a variety of vision tasks such as object detection, instance segmentation, and pose estimation. In contrast to the object-centered images in ImageNet, COCO images contain natural scenes with complex backgrounds, multiple objects of different categories, and different resolutions. The dataset includes 200k images from 80 object classes, and all object instances are annotated with detailed segmentation masks. For the human class, the dataset includes 150k human instances, each annotated with 17 different keypoints, making it suitable for the human pose estimation task.

\subsubsection{Evaluation metrics}
For classification, we report the widely used Top-1 and Top-5 accuracy as the evaluation metric, while for object detection, instance segmentation, and human pose estimation tasks, we adopt the widely used mean average precision (mAP) over all classes/keypoints as the primary evaluation metric. The mAP value is obtained by averaging the average precision (AP) scores calculated given a series of thresholds ranging from 0.5 to 0.95 with a step of 0.05. For pose estimation, we use the object keypoint similarity (OKS), to determine the threshold. For semantic segmentation tasks, we use the intersection over union (IoU) as the evaluation metric.

\subsubsection{Implementation details}
In this study, we employ QA in both plain and hierarchical vision transformers. We adopt the same model specifications as ViT \cite{vit} and Swin Transformer \cite{liu2021swin} for QFormer$_p$ and QFormer$_h$, respectively. If not specified, the input image resolution is set to $224 \times 224$ for image classification. For hierarchical architectures, we follow the hyper-parameter settings of Swin Transformer to train QFormer$_h$ in a supervised manner. For plain architectures, we adopt the MAE~\cite{he2021masked} pre-training strategy to provide a good initialization for QFormer$_p$, which has been proven effective in mitigating over-fitting in plain ViTs. For different downstream tasks, we used the ImageNet-1k pre-trained weights for initialization. It is worth noting that we only incorporate the proposed QA during the fine-tuning phase of QFormer$_p$ to reuse the pre-trained weights of plain ViTs and reduce the training cost, although incorporating quadrangle attention in pre-training may result in better performance.

\subsection{Plain models}
\subsubsection{Image classification}
\noindent \textbf{Settings.} For the plain models, we adopt the fine-tuning practice proposed in \cite{he2021masked}, which involves pre-training the model on the ImageNet-1K dataset for 1,600 epochs using the ImageNet-1K training set, followed by supervised fine-tuning. We make several modifications to the backbone network, including introducing relative position encodings in each attention layer to incorporate position information, using average pooling of all visual tokens instead of the class token for the final prediction, and implementing window attention with a window size of 7 \cite{liu2021swin} and a patch size of 16 \cite{vit}, following the common practice in \cite{li2022exploring}. The hyper-parameter for the proposed regularization term is set to 1, and all other parameters follow the same settings as in \cite{he2021masked}. We employ the AdamW optimizer with a weight decay of 0.05 and a layer-wise learning rate decay of 0.75. The model was trained for 100 epochs, with the first 5 epochs for warm-up.

\begin{table}[htbp]
  \centering
  \caption{Ablation study results of each basic transformations on the QFormer$_p$-B model. }
    \setlength{\tabcolsep}{0.007\linewidth}\begin{tabular}{cccc|cc}
    \hline
    \multicolumn{4}{c|}{Transformations} & \multicolumn{2}{c}{ImageNet} \\
    \hline
    scale and shift $(T_s\cdot T_t)$ & shear $(T_h)$ & rotation $(T_r)$ & projection $(T_p)$ & Top-1 & Top-5 \\
    \hline
          &       &       &       & 81.2  & 95.5 \\
    \hline
    \checkmark     &       &       &       & 82.3  & 96.1 \\
    \checkmark     & \checkmark     &       &       & 82.5  & 96.1 \\
    \checkmark     & \checkmark     & \checkmark     &       & 82.6  & 96.2 \\
    \checkmark     & \checkmark     & \checkmark     & \checkmark     & 82.9  & 96.3 \\
    \hline
    \end{tabular}%
  \label{tab:quadra ablation}%
\end{table}%

\begin{table}[htbp]
  \centering
  \caption{Comparison between window attention and quadrangle attention, where full attention is placed evenly in the ViT-B/16 model following ViTDet \cite{li2022exploring}.}
    \begin{tabular}{ccc|ccc}
    \hline
    \multicolumn{3}{c|}{Attention number} & \multicolumn{3}{c}{Accuracy} \\
    Window & Quadrangle & Full  & Top-1 & Gain ($\Delta$) & Top-5  \\
    \hline
    \checkmark &       & \multirow{2}[1]{*}{0} & 81.2 &  & 95.5 \\
          & \checkmark &       & 82.9 & +1.7  & 96.3\\
    \hline
    \checkmark &       & \multirow{2}[0]{*}{2} & 82.5 &  & 96.1 \\
          & \checkmark &       & 83.0 & +0.5  & 96.3 \\
    \hline
    \checkmark &       & \multirow{2}[0]{*}{3} & 82.6 &  & 96.2 \\
          & \checkmark &       & 83.1 & +0.5  & 96.3 \\
    \hline
    \checkmark &       & \multirow{2}[0]{*}{4} & 82.8 & & 96.3 \\
          & \checkmark &       & 83.2 & +0.4 & 96.4 \\
    \hline
    \checkmark &       & \multirow{2}[1]{*}{6} & 82.9 &  & 96.4 \\
          & \checkmark &       & 83.3 & +0.4 & 96.6 \\
    \hline
    \end{tabular}%
  \label{tab:ablation on qa num}%
\end{table}%

\begin{table}[htbp]
  \centering
  \caption{Comparison of using different attention methods in ViT-B/16 on ImageNet-1k. Quadrangle* refers to directly predicting the transformation while Quadrangle denotes the composition of basic transformation as depicted in Eq.~\eqref{eq:basic composition}.}
    \begin{tabular}{c|c|ccc}
    \hline
          & \multicolumn{1}{c|}{Accuracy} & \multicolumn{3}{c}{Traing cost} \\
\cline{2-5}          & Top-1& Epoch time & FLOPs (G) & Memory (MB) \\
    \hline
    Window & 81.2  & 6:20  & 16.9  & 13,045 \\
    Shift window & 82.0 & 7:48  & 16.9  & 13,238 \\
    Quadrangle* & 82.7 & 8:10 &  16.9 & 15,320 \\
    Quadrangle & 82.9 & 8:19  & 16.9  & 15,331 \\
    \hline
    \end{tabular}%
  \label{tab:classification window comparison}%
\end{table}%

\begin{table}[htbp]
  \centering
  \caption{Hyper-parameter study of $\lambda$ for the regularization.}
    \begin{tabular}{c|ccccccc}
    \hline
    $\lambda$ & 0   & 0.0001 & 0.001 & 0.01 & 0.1 & 1     & 10 \\
    \hline
    Top-1 & diverged & 82.6  & 82.7  & 82.8  & 82.9  & 82.9  & 82.7 \\
    \hline
    \end{tabular}%
  \label{tab:hyper parameter lambda}%
\end{table}%

\begin{table*}[htbp]
  \centering
  \caption{Object detection results on MS COCO~\cite{lin2014microsoft} with the Mask RCNN detector~\cite{he2017mask}. Full attention is used every 1/4 attention layers. The results of bounding box mAP$^{bb}$ and instance mask mAP$^{mk}$ are reported, respectively.}
    \begin{tabular}{c|cccccc|cccccc|cc}
    \hline
    Model & mAP$^{bb}$ & AP$^{bb}_{50}$ & AP$^{bb}_{75}$ & AP$^{bb}_s$ & AP$^{bb}_m$ & AP$^{bb}_l$ & mAP$^{mk}$ & AP$^{mk}_{50}$ & AP$^{mk}_{75}$ & AP$^{mk}_s$ & AP$^{mk}_m$ & AP$^{mk}_l$ & FLOPs & Params \\
    \hline
    ViTDet-B \cite{li2022exploring} & 51.6  & 72.3  & 57.1  & 35.6  & 55.3  & 66.5  & 45.9  & 69.4  & 50.0 & 27.8  & 48.7  & 63.7  & 0.8T  & 111M \\
    QFormer$_p$-B & 52.3  & 73.0  & 57.7  & 36.2  & 55.5  & 67.4  & 46.6  & 70.4  & 50.9  & 27.8  & 49.3  & 65.3  & 0.8T  & 111M \\
    \hline
    \end{tabular}%
  \label{tab:plain detection}%
\end{table*}%

\begin{table*}[htbp]
  \centering
  \caption{The semantic segmentation results on ADE20k~\cite{zhou2017scene} with UPerNet \cite{xiao2018unified} as the segmentation head. * denotes the results obtained by the multi-scale test.}
    \begin{tabular}{c|ccc|ccc|ccc|ccc}
    \hline
          & \multicolumn{6}{c|}{Image size 512}           & \multicolumn{6}{c}{Image size 640} \\
\cline{2-13}          & mIoU  & mAcc  & aAcc  & mIoU* & mAcc* & aAcc* & mIoU  & mAcc  & aAcc  & mIoU* & mAcc* & aAcc* \\
    \hline
    ViT-B + window attn & 39.7 & 50.3 & 79.1 & 41.8 & 51.0 & 80.8 & 40.2  & 50.7 & 79.4 & 41.5 & 50.7 & 80.6 \\
    ViT-B + shifted window attn & 41.6 & 53.3 & 79.9 & 43.6 & 53.9 & 81.4 & 42.3 & 53.6 & 80.6 & 43.5 & 53.4 & 81.5 \\
    QFormer$_p$-B & 43.6 & 55.2 & 80.6 & 45.0 & 55.4 & 82.1 & 44.9 & 56.1 & 81.7 & 46.0 & 56.0 & 82.6 \\
    \hline
    \end{tabular}%
  \label{tab:plain segmentation}%
\end{table*}%

\noindent \textbf{Ablation study.} In order to evaluate the effectiveness of the flexible window design in QA, we conduct an ablation study using QFormer$_p$-B. We gradually add basic transformations in QA, including scale and shift, shear, rotation, and projection transformations. The results are presented in Table~\ref{tab:quadra ablation}, where $\checkmark$ indicates that the corresponding transformation is used. The first row denotes the default setting, \ie, using the fixed-size windows. We observe that the representation ability of the model improves continuously with using more transformations, with the Top-1 accuracy increasing from 81.2 to 82.9 (+1.7). This improvement is attributed to the ability of the generated flexible quadrangles to capture richer context and better handle objects of different sizes, orientations, and shapes. Furthermore, this flexible window configuration enables the network to capture long-range dependency better, thereby unleashing the potential of self-attention.

Another way to address the limited attention distance problem in window attention is to combine it with the vanilla full attention, which allows each token to attend to all other tokens, thus enabling the modeling of long-range dependency \cite{li2022exploring}. ViTDet \cite{li2022exploring} interleaves window attention and full attention layers to enhance the model's ability. We investigate the effectiveness of QA in this setting. Specifically, we vary the number of full attention layers in the network, which are evenly distributed throughout the network, and evaluate the performance of using either fixed-size window attention or quadrangle attention in other transformer layers. The QFormer$_p$-B backbone with 12 layers is used for the experiments. The results are presented in Table~\ref{tab:ablation on qa num}, where the third column indicates the number of full attention layers. It is surprising that even with the usage of full attention, QA continuously outperforms window attention in all settings. For example, with no full attention involved, QA brings the most significant performance gain, \ie, from 81.2 to 82.9 Top-1 accuracy. This is due to the better ability of QA to model long-range dependency and promote cross-window information exchange. It can also explain the fewer gains brought by the full attention for QA, compared with the window attention. It is noteworthy that the model using six full attention layers and window attention (\ie, the first row in the last group) only obtains comparable performance with the model only using QA (\ie, the second row in the first group). Since the QA has much fewer computations than the full attention, the results validate the superiority of the proposed QA.

To further validate the effectiveness of QA, we compare it with the window attention as well as its shifted window version \cite{liu2021swin} using ViT-B/16 as the backbone. The results in Table~\ref{tab:classification window comparison} show that the proposed QA achieves the best performance with a Top-1 accuracy of 82.9 and even outperforms the shifted window attention by a gain of 0.9, showing that flexible quadrangles in QA not only promote cross-window information exchange as shifted window attention but also help to extract rich context and learn better feature representation. Moreover, the proposed decomposed formulation, as illustrated in Eq.~\eqref{eq:basic composition}, yields a 0.2 performance improvement over directly predicting the transformation matrix. This highlights the effectiveness of the composition of basic transformations. The training cost of QA is similar to shifted window attention in terms of training time per epoch, overall FLOPs, and memory footprint, indicating that the proposed QA makes a good trade-off between accuracy and computational cost.

We conduct an analysis of the hyper-parameter $\lambda$ in the proposed regularization term, and the results are presented in Table~\ref{tab:hyper parameter lambda}, where $\lambda$ ranges from 0 to 10. All results are obtained using QFormer$_p$-B. When $\lambda=0$, the regularization term is not applied during training. In this case, the quadrangle covers too many invalid regions (\ie, regions outside the feature maps), and the gradients from these regions are always zero, resulting in model divergence. By incorporating the proposed regularization term to penalize unreasonable attention regions, we stabilize the training process. We sweep the value of $\lambda$, and the best performance can be achieved in the range of $0.1\sim1$.

\subsubsection{Object detection and instance segmentation}
\noindent \textbf{Settings.} The object detection model is built and trained following ViTDet \cite{li2022exploring}. Specifically, features are extracted from the last layer of the backbone, and a feature pyramid is constructed by downsampling and upsampling the extracted features. To better promote long-term dependency modeling, four full attention layers are uniformly placed in the backbone, while the other attention layers adopt the proposed QA. The official MAE pre-trained weights for the backbone are utilized, and the entire model is fine-tuned for 100 epochs on the MS COCO dataset. During training, the drop path rate is set to 0.2, and hyper-parameters from ViTDet are adopted, such as the AdamW optimizer with an initial learning rate of 0.0001, weight decay of 0.1, and layer-wise learning rate decay of 0.7. The input image resolution is set to 1024$\times$1024. The performance of bounding box object detection (mAP$^{bb}$) and instance segmentation (mAP$^{mk}$) is evaluated and reported.

\noindent\textbf{Results.} Table~\ref{tab:plain detection} presents the results on the MS COCO validation set. QFormer$_p$-B and ViTDet-B differ only in the use of the proposed QA or window attention. The QFormer$_p$-B model outperforms its baseline ViTDet-B model by a gain of 0.7 mAP$^{bb}$ and 0.7 mAP$^{mk}$ while maintaining the model size and computational efficiency, as indicated by FLOPs and parameters. This considerable improvement demonstrates the effectiveness of QA in helping the detector learn better object representation. Moreover, QFormer$_p$-B demonstrates superior performance in $AP_s$ and $AP_l$, exhibiting an improvement of 0.6 AP$^{bb}_s$, 0.9 AP$^{bb}_l$, and 1.6 AP$^{mk}_l$. This outcome suggests that QA can handle objects of different scales due to its flexible design of window configuration.

\begin{table*}[htbp]
  \centering
  \caption{Human pose estimation results on the MS COCO~\cite{lin2014microsoft} validation set. * denotes that FP16 is used due to the limit of GPU memory. When `Full' is marked, we use full attention every 1/4 layer following ViTDet~\cite{li2022exploring}.}
    \begin{tabular}{cccc|cc|cccc|c}
    \hline
    Full  & Window & Shift & QA    & Training Memory (M) & GFLOPs & AP    & AP$_{50}$  & AR    & AR$_{50}$  & Training time (min/epoch) \\
    \hline
          & \checkmark &       &       & 21,161 & 66.3 & 66.4  & 87.7  & 72.9  & 91.9  & 3:50 \\
         & \checkmark & \checkmark &       & 21,161 & 66.3 & 76.4  & 90.9  & 81.6  & 94.5  & 4:25 \\
          &       &       & \checkmark & 24,378 &  66.4 & 77.0   & 90.9  & 82.0    & 94.7  & 4:36 \\
    \hline
    \checkmark & \checkmark &       &       & 28,594 & 69.9 & 76.9  & 90.8  & 82.1  & 94.7  & 4:20 \\
    \checkmark &       &       & \checkmark & 32,641 & 70.0 & 77.2 & 90.9 & 82.2  & 94.7 & 4:40 \\
    \checkmark &       &       &       & 36,141* & 76.6 & 77.4  & 91.0    & 82.4  & 94.9  & 4:10 \\
    \hline
    \end{tabular}%
  \label{tab:pose}%
\end{table*}%

\subsubsection{Semantic segmentation}
\noindent\textbf{Settings.} We evaluate our QFormer for semantic segmentation on the ADE20k dataset. In contrast to the detection task, where only the last layer is used for feature extraction, features are collected from every 1/4 layer of the ViT-B backbone for the segmentation task. A feature pyramid is constructed for the segmentation head using simple deconvolution and downsampling layers. The MAE pre-trained weights are used to initialize the backbone, and the whole model is fine-tuned for 160k iterations. The UPerNet \cite{xiao2018unified} is adopted as the segmentation head, following the common practice \cite{swinv2,he2021masked}, and the default training setting in MMSegmentation~\cite{contributors2020openmmlab} is adopted. For example, the AdamW optimizer is used with an initial learning rate of 6e-5. A polynomial scheduler is employed to adjust the learning rate, while weight decay and layer-wise learning rate decay are set to 0.01 and 0.9, respectively. Two training settings with different input sizes, namely 512$\times$512 and 640$\times$640, are used to compare the models' performance. All the models are trained using 8 A100 GPUs with a total batch size of 16. The results are reported in terms of mIoU and accuracy, for both single-scale and multi-scale test settings.

\noindent\textbf{Results.} Table~\ref{tab:plain segmentation} presents the results. The first two rows denote the results of using fixed window attention or shifted window attention \cite{swinv2} in ViT-B, respectively. The last row reports the results of using the proposed QA in ViT-B, \ie, QFormer$_p$-B. The superior performance of QA can be observed in comparison to the other two variants, especially for this dense prediction task with various objects. Specifically, QFormer$_p$-B achieves 43.6 mIoU for 512$\times$512 images and 44.9 mIoU for 640$\times$640 images, which outperforms the shifted window attention by 2.0 and 2.6 mIoU, respectively. This improvement is due to the QA's flexible design that learns adaptive window configurations to extract rich context and better feature representation. Moreover, when comparing models trained with different input resolutions, it can be observed that the models with the default or shifted window attention gain less or even no benefit from larger images. The default window attention yields inferior performance in this setting, as it lacks the ability to model global context. The shifted window attention helps promote cross-window information exchange and improves performance slightly. In contrast, QFormer$_p$-B can adapt to objects of different sizes and effectively exploit the context information in larger images, resulting in better segmentation performance. For instance, QFormer$_p$-B gains 1.4 mIoU and 1.1 mIoU* when increasing the input resolution from 512 to 640.

\subsubsection{Human pose estimation}
\noindent\textbf{Settings.} We follow the practice in ViTPose \cite{xu2022vitpose} to evaluate the effectiveness of QA for the human pose estimation task. The top-down pipeline is employed, where the images of human instances are used as input, extracted based on the detection results from SimpleBaseline \cite{xiao2018simple}. Post-processing is carried out using Udp~\cite{huang2020devil}. We use the official MAE pre-trained model to initialize the ViT-B backbone and the default training settings in MMPose, \ie, an input image size of 256$\times$192 and a learning rate of 5e-4. The model is optimized by the AdamW optimizer and trained for 210 epochs. The learning rate is decreased by 10$\times$ at the 170th and 200th epochs, with the layer-wise decay set to 0.75. Larger feature size is used by adjusting the convolutional stride in the patch embedding layer, changing the downsampling ratio from 1/16 to 1/8. The performance of different attention methods, including the vanilla full attention (`Full'), fixed window attention (`Window'), shifted window attention (`Shift'), and the proposed QA, are reported on the MS COCO dataset~\cite{lin2014microsoft}.

\begin{table*}[h]
  \centering
  \caption{Image classification results on ImageNet~\cite{deng2009imagenet}. `Input Size' denotes the size of input images used for training and testing.}
    \begin{tabular}{l|ccc|cc|c}
    \hline
    \multicolumn{1}{c|}{\multirow{2}[2]{*}{Model}} & Params & FLOPs & Input & \multicolumn{2}{c|}{ImageNet~\cite{deng2009imagenet}} & Real~\cite{beyer2020we} \\
          & (M)   & (G)   & Size  & Top-1 & Top-5 & Top-1 \\
    \hline
    DeiT-S~\cite{touvron2020training} & 22  & 4.6   & 224   & 81.2  & 95.4  & 86.8  \\
    PVT-S~\cite{wang2021pyramid} & 25  & 3.8   & 224   & 79.8  & - & -  \\
    ViL-S~\cite{zhang2021multi} & 25 & 4.9 & 224 & 82.4 & - & - \\
    PiT-S~\cite{heo2021pit} & 24  & 4.8   & 224   & 80.9  &   -    & - \\
    TNT-S~\cite{han2021transformer} & 24  & 5.2   & 224   & 81.3  & 95.6  & - \\
    MSG-T~\cite{fang2021msg} & 25 & 3.8 & 224 & 82.4 & - & - \\
    Twins-PCPVT-S~\cite{chu2021twins} & 24  & 3.8   & 224   & 81.2  &    -   &  - \\
    Twins-SVT-S~\cite{chu2021twins} & 24  & 2.9   & 224   & 81.7  & -      & - \\
    T2T-ViT-14~\cite{yuan2021tokens} & 22  & 5.2   & 224   & 81.5  & 95.7  & 86.8  \\
    Swin-T~\cite{liu2021swin} & 29  & 4.5   & 224   & 81.2  &    -   & - \\
    Focal-T~\cite{yang2021focal} & 29 & 4.9 & 224 & 82.2 & 95.9 & - \\
    DW-T~\cite{ren2022beyond} & 30 & 5.2 & 224 & 82.0 & - & - \\
    {QFormer$_h$-T} & {29} & {4.6} & {224} & {82.5} & {96.2}  & {87.5} \\
    \hline
    Swin-T~\cite{liu2021swin} & 29  &  12.5  & 320   & 81.6  &   &   \\
    {QFormer$_h$-T} & {29} & 13.0 & 320 & 83.0 & 96.3 &  \\
    Swin-T~\cite{liu2021swin} & 29  &  14.2  & 384   & 82.0  &   &  \\
    {QFormer$_h$-T} & {29} & {14.9} & {384} & {83.2} & {96.5}  & {88.0} \\
    \hline
    PiT-B~\cite{heo2021pit} & 74  & 12.5  & 224   & 82.0  &   -    & - \\
    TNT-B~\cite{han2021transformer} & 66  & 14.1  & 224   & 82.8  & 96.3  & - \\
    ViL-B~\cite{zhang2021multi} & 56 & 13.4 & 224 & 83.7 & - & - \\
    MSG-S~\cite{fang2021msg} & 56 & 8.4 & 224 & 83.4 & -  & - \\
    PVTv2-B5~\cite{wang2021pvtv2} & 82  & 11.8  & 224   & 83.8  &  -     & - \\
    HRFormer-B \cite{yuan2021hrformer} & 50 & 13.7 & 224 & 82.8 & - & - \\
    Swin-S~\cite{liu2021swin} & 50  & 8.7   & 224   & 83.2  &  96.2  &  -\\
    Focal-S~\cite{yang2021focal} & 	51 &  9.4  & 224 & 83.5 &  96.2 & - \\
    CrossFormer-B~\cite{wang2021crossformer} & 52 & 9.2 & 224 & 83.4 &  \\
    {QFormer$_h$-S} & {51} & {8.9} & {224} & {84.0} &  {96.8} & {88.6} \\
    \hline
    Swin-B~\cite{liu2021swin} & 88  & 15.4  & 224   & 83.4  &  96.5 & 88.0 \\
    DW-B \cite{ren2022beyond} & 91  &  17.0  & 224   & 83.8  & - & - \\
    Focal-B~\cite{yang2021focal} & 90  & 16.0  & 224   & 83.8  &  96.5  & - \\
    {QFormer$_h$-B} & {90} & {15.7} & {224} & {84.1} &  {96.8} & {88.7} \\
    \hline
    
    \end{tabular}%
  \label{tab:Classification}%
\end{table*}%

\noindent\textbf{Results.} As shown in Table~\ref{tab:pose}, the model with window attention only, represented by the first row, demonstrates inferior performance due to its poor ability to extract global context information. Both the shifted window attention and our QA promote cross-window information exchange and show improved performance, where QA obtained larger gains in most metrics. The results suggest that the learned quadrangles of various shapes can adapt well to humans with diverse poses and learn better feature representation. Furthermore, when comparing the 3rd and 4th rows, we find that using QA alone has already outperformed the model of using both full and window attention but with less GPU memory footprint and fewer FLOPs. When using full attention and QA together, the performance reaches 77.2 AP (the second-last row) and is comparable to the full attention setting, while requiring much less memory footprint and FLOPs.

\begin{table*}[htbp]
  \centering
  \caption{Object detection results on MS COCO~\cite{lin2014microsoft} with the Mask RCNN detector~\cite{he2017mask}.}
    \begin{tabular}{l|c|ccc|ccc|ccc|ccc}
    \hline
          & {Params} & \multicolumn{6}{c|}{Mask RCNN 1x}             & \multicolumn{6}{c}{Mask RCNN 3x} \\
          & (M) & mAP$^{bb}$ & AP$_{50}^{bb}$ & AP$_{75}^{bb}$ & mAP$^{mk}$ & AP$_{50}^{mk}$ & AP$_{75}^{mk}$ & mAP$^{bb}$ & AP$_{50}^{bb}$ & AP$_{75}^{bb}$ & mAP$^{mk}$ & AP$_{50}^{mk}$ & AP$_{75}^{mk}$ \\
    \hline
    ResNet50~\cite{he2016deep} & 44  & 38.6  & 59.5  & 42.1  & 35.2  & 56.3  & 37.5  & 40.8 &  61.2 &  44.4 & 37.0 &    58.4 & 39.3 \\
    ViL-S~\cite{yu2016multi} & 45  & 44.9  & 67.1  & 49.3  & 41.0  & 64.2  & 44.1  & 47.1  & 68.7  & 51.5  & 42.7  & 65.9  & 46.2  \\
    PVT-M~\cite{wang2021pyramid} & 64  & 42.0  & 64.4  & 45.6  & 39.0  & 61.6  & 42.1  & -     & -     & -     & -     & -     & - \\
    PVT-L~\cite{wang2021pyramid} & 81  & 42.9  & 65.0  & 46.6  & 39.5  & 61.9  & 42.5  &       &       &       &       &       &  \\
    PVTv2-B2~\cite{wang2021pvtv2} & 45  & 45.3  & 67.1  & 49.6  & 41.2  & 64.2  & 44.4  & -     & -     & -     & -     & -     & - \\
    CMT-S~\cite{guo2021cmt} & 45  & 44.6  & 66.8  & 48.9  & 40.7  & 63.9  & 43.4  & -     & -     & -     & -     & -     & - \\
    RegionViT-S~\cite{chen2021regionvit} & 50  & 42.5  & -     & -     & 39.5  & -     & -     & 46.3  & -     & -     & 42.3  & -     & - \\
    XCiT-S12/16~\cite{el2021xcit} & 44  & -     & -     & -     & -     & -     & -     & 45.3  & 67.0  & 49.5  & 40.8  & 64.0  & 43.8  \\
    DPT-M~\cite{chen2021dpt} & 66  & 43.8  & 66.2  & 48.3  & 40.3  & 63.1  & 43.4  & 44.3  & 65.6  & 48.8  & 40.7  & 63.1  & 44.1  \\
    ResT-Base~\cite{zhang2021rest} & 50  & 41.6  & 64.9  & 45.1  & 38.7  & 61.6  & 41.4  & -     & -     & -     & -     & -     & - \\
    Shuffle-T~\cite{huang2021shuffle} & 48 & - & - & - & - & - & - & 46.8 & 68.9 & 51.5 & 42.3 & 66.0 & 45.6 \\
    Swin-T~\cite{liu2021swin} & 48  & 43.7  & 66.6 & 47.7 & 39.8  & 63.3 & 42.7 & 46.0  & 68.1  & 50.3 & 41.6  & 65.1  & 44.9 \\
    DW-T~\cite{ren2022beyond} & 49 & -     & -     & -     & -     & -     & -     & 46.7 & 69.1 & 51.4 & 42.4 & 66.2 & 45.6 \\
    Focal-T~\cite{yang2021focal} & 49 & 44.8 & - & - & 41.0 & - & - & 47.2 & 69.4 & 51.9 & 42.7 & 66.5 & 45.9 \\
    DAT-T \cite{xia2022vision} & 48 & 44.4 & 67.6 & 48.5 & 40.4 & 64.2 & 43.1 & 47.1 & 69.2 & 51.6 & 42.4 & 66.1 & 45.5 \\
    {QFormer$_h$-T} & {49}  & {45.9}  & {68.5}  & {50.3} & {41.5}  & {65.2} & {44.6} & {47.5}  & {69.6} & {52.1} & {42.7}  & {66.4} & {46.1} \\
    \hline
    \end{tabular}%
  \label{tab:hierarchical MaskRCNN}%
\end{table*}%

\begin{table*}[htbp]
  \centering
  \caption{Object detection results on MS COCO~\cite{lin2014microsoft} with the Mask RCNN detector~\cite{he2017mask}.}
    \begin{tabular}{c|c|ccc|ccc}
    \hline
          & Params & \multicolumn{6}{c}{Mask RCNN 3x} \\
\cline{3-8}          &  (M)  & mAP$^{bb}$ & AP$^{bb}_{50}$ & AP$^{bb}_{75}$ & mAP$^{mk}$ & AP$^{mk}_{50}$ & AP$^{mk}_{75}$ \\
    \hline
    PVT-L \cite{wang2021pyramid} & 81    & 44.5  & 66.0  & 48.3  & 40.7  & 63.4  & 43.7 \\
    ViL-B \cite{zhang2021multi} & 76    & 45.7  & 67.2  & 49.9  & 41.3  & 64.4  & 44.5 \\
    Swin-S \cite{liu2021swin} & 69    & 48.5  & 70.2  & 53.5  & 43.3  & 67.3  & 46.6 \\
    Focal-S \cite{yang2021focal} & 71    & 48.8  & 70.5  & 53.6  & 43.8  & 67.7  & 47.2 \\
    DAT-S \cite{xia2022vision} & 69    & 49.0   & 70.9  & 53.8  & 44.0  & 68.0    & 47.5 \\
    Swin-B \cite{liu2021swin} & 107 & 48.5 & 69.8 & 53.2 & 43.4 & 66.8 & 46.9 \\
    DW-B \cite{ren2022beyond} & 111 & 49.2 & 70.6 & 54.0 & 44.0 & 68.0 & 47.7 \\
    {QFormer$_h$-S} & {70} & {49.5} & {71.2} & {54.2} & {44.2} & {68.3} & {47.6} \\
    \hline
    \end{tabular}%
  \label{tab:hierarchical Mask RCNN larger}%
\end{table*}%

\subsection{Hierarchical models}

\subsubsection{Image classification}
\noindent \textbf{Settings.} During training, we utilize the AdamW optimizer \cite{loshchilov2018decoupled} with a cosine learning rate scheduler. The training process consists of 300 epochs on the ImageNet-1k \cite{deng2009imagenet} training set, and we also use linear warm-up for the first 20 epochs. The initial learning rate is set to 0.001 with a batch size of 1024. We apply various data augmentations, including random cropping, auto-augmentation \cite{autoaug}, CutMix \cite{yun2019cutmix}, MixUp \cite{zhang2017mixup}, and random erasing. Furthermore, label smoothing with a weight of 0.1 is used. For training with input size larger than 224$\times$224, we fine-tune the model pre-trained with an input size of 224$\times$224, for 30 epochs using the same training settings but without the warm-up phase.

\noindent \textbf{Results.} We evaluate different models on the ImageNet-1k~\cite{deng2009imagenet} validation set. As shown in Table~\ref{tab:Classification}, the proposed QFormer$_h$-T outperforms its counterpart Swin-T (with the shifted window attention) by a gain of 1.3\% Top-1 accuracy, \ie,  from 81.2\% to 82.5\%. This result suggests that QA can capture useful context information more effectively by learning the appropriate window configurations and thus attending to far-away but relevant tokens outside the default windows. Moreover, QFormer$_h$-T achieves superior results over MSG-T~\cite{fang2021msg} and Focal-T~\cite{yang2021focal}, which utilize additional messenger tokens for across-window information exchange, demonstrating that our QA can enable sufficient information exchange across windows without hand-crafted designs, making it simple and efficient. When increasing the model size, our QFormer$_h$-S outperforms Swin-S by 1.0\% accuracy and it even surpasses the larger Swin-B and Focal-B by 0.6\% and 0.2\% accuracy, respectively, while having significantly fewer parameters and FLOPs. The performance gains are not diminished when further scaling the QFormer to larger sizes, \eg, QFormer$_h$-B achieves an even better performance of 84.1\% accuracy, although the gain over its smaller version QFormer$_h$-S is marginal. We suspect that the performance of large models may be limited by the small input size.

\begin{table*}[htbp]
  \centering
  \caption{Object detection results on MS COCO~\cite{lin2014microsoft} with the Cascade Mask RCNN detector~\cite{cai2019cascade}.}
    \setlength{\tabcolsep}{0.01\linewidth}{\begin{tabular}{l|c|ccc|ccc|ccc|ccc}
    \hline
          & Params & \multicolumn{6}{c|}{Cascade RCNN 1x}     & \multicolumn{6}{c}{Cascade RCNN 3x} \\
          & (M) & mAP$^{bb}$ & AP$_{50}^{bb}$ & AP$_{75}^{bb}$ & mAP$^{mk}$ & AP$_{50}^{mk}$ & AP$_{75}^{mk}$ & mAP$^{bb}$ & AP$_{50}^{bb}$ & AP$_{75}^{bb}$ & mAP$^{mk}$ & AP$_{50}^{mk}$ & AP$_{75}^{mk}$ \\
    \hline
    R50 \cite{he2016deep} & 82    & 44.3  & 62.7  & 48.4  & 38.3  & 59.7  & 41.2  & 46.3  & 64.3  & 50.5  & 40.1  & 61.7  & 43.4 \\
    Swin-T \cite{liu2021swin} & 86    & 48.1  & 67.1  & 52.2  & 41.7  & 64.4  & 45.0  & 50.2  & 69.2  & 54.7  & 43.7  & 66.6  & 47.3 \\
    DAT-T \cite{xia2022vision} & 86 & 49.1 & 68.2 & 52.9 & 42.5 & 65.4 & 45.8 & 51.3 & 70.1 & 55.8 & 44.5 & 67.5 & 48.1 \\
    {QFormer$_h$-T} & {87} & {49.8} & {69.0} & {54.1} & {43.0} & {66.1} & {46.7} & {51.4} & {70.1} & {55.8} & {44.7} & {67.6} & {48.2} \\
    \hline
    \end{tabular}}%
  \label{tab:hierarchical Cascade RCNN}%
\end{table*}%

To investigate the impact of image size, we compare the performance of Swin and our QFormer$_h$ under different settings as shown in Figure~\ref{fig:opening}(c). Due to the limit of GPU memory, we only conduct experiments on Swin-T and our QFormer$_h$-T. When increasing the input image size from 224 $\times$ 224 to 320 $\times$ 320 and 384 $\times$ 384, we observe that the proposed QA continually helps deliver better classification performance, which outperforms the (shifted) window attention significantly under all the settings. It is noteworthy that QFormer$_h$-T achieves 82.5\% accuracy with the image size of 224 $\times$ 224, even outperforming Swin-T at 384 $\times$ 384 by a gain of 0.5. The results suggest that window attention that limits the attention area to a fixed-size square makes it ineffective in handling objects of different sizes, while the proposed quadrangle attention is very flexible and learns adaptive windows from the data directly.

\begin{table*}[htbp]
  \centering
  \caption{Object detection results on MS COCO~\cite{lin2014microsoft} with the Cascade Mask RCNN detector~\cite{cai2019cascade}.}
    \begin{tabular}{c|c|cccccc|cccccc}
    \hline
          & Params & \multicolumn{12}{c}{Cascade RCNN 3x} \\
\cline{3-14}          &  (M)  & mAP$^{bb}$ & AP$^{bb}_{50}$ & AP$^{bb}_{75}$ & AP$^{bb}_{s}$ &  AP$^{bb}_{m}$  & AP$^{bb}_{l}$  & mAP$^{mk}$ & AP$^{mk}_{50}$ & AP$^{mk}_{75}$ & AP$^{mk}_{s}$ & AP$^{mk}_{m}$ & AP$^{mk}_{l}$ \\
    \hline
    X101-32 \cite{xie2017aggregated} & 101   & 48.1  & 66.5  & 52.4  & -     & -     & -     & 41.6  & 63.9  & 45.2  & -     & -     & - \\
    Swin-S \cite{liu2021swin} & 107   & 51.9  & 70.7  & 56.3  & 35.2  & 55.7  & 67.7  & 45.0    & 68.2  & 48.8  & 28.8  & 48.7  & 60.6 \\
    Shuffle-S \cite{huang2021shuffle} & 107   & 51.9  & 70.9  & 56.4  & -     & -     & -     & 44.9  & 67.8  & 48.6  & -     & -     & - \\
    MSG-S \cite{fang2021msg} & 113 & 52.5 & 71.1 & 57.2 & -     & -     & - & 45.5 & 68.4 & 49.5 & -     & -     & -\\
    QFormer$_h$-S & 108   & 52.8  & 71.5  & 57.4  & 35.9  & 56.6  & 68.5  & 45.7  & 69.1  & 50.1  & 29.2  & 49.6  & 61.7 \\
    \hline
    Swin-B \cite{liu2021swin} & 145   & 51.9  & 70.5  & 56.4  & 35.4  & 55.2  & 67.4  & 45.0    & 68.1  & 48.9  & 28.9  & 48.3  & 60.4 \\
    QFormer$_h$-B & 147   & 52.9  & 71.6  & 57.6  & 36.3  & 56.5  & 68.5  & 45.9  & 69.2  & 49.8  & 29.7  & 49.4  & 61.5 \\
    \hline
    \end{tabular}%
  \label{tab:hierarchical Cascade RCNN larger}%
\end{table*}%

\subsubsection{Object detection and instance segmentation}
\textbf{Settings.} We evaluate QA in hierarchical models for the object detection and instance segmentation tasks on the MS COCO~\cite{lin2014microsoft} dataset. We utilize backbones pre-trained on ImageNet with an input size of 224 $\times$ 224 and two popular object detection frameworks, namely Mask RCNN~\cite{he2017mask} and Cascade RCNN~\cite{cai2018cascade,cai2019cascade}. We train and evaluate all the models using the common practices according to mmdetection~\cite{mmdetection}. Specifically, we use the AdamW optimizer and a batch size of 16 for multi-scale training. The initial learning rate and weight decay are set to 0.0001 and 0.05, respectively. We train the models under both 1$\times$ (12 epochs) and 3$\times$ (36 epochs) schedules.

\begin{table}[htbp]
  \centering
  \caption{Semantic segmentation results on ADE20k~\cite{zhou2017scene} with UPerNet~\cite{xiao2018unified}. mIoU* denotes the result of multi-scale testing.}
    \begin{tabular}{c|ccc|c|c}
    \hline
          & mIoU  & mAcc  & aAcc  & mIoU* & Params \\
    \hline
    ResNet-50 \cite{he2016deep} & 42.1 & 53.0 & 80.0 & 42.8 & 67M \\
    Swin-T \cite{liu2021swin} & 44.5  & 55.6  & 81.1  & 45.8  & 60M \\
    DAT-T \cite{xia2022vision} & 45.5 & 57.9  & -  & 46.4  & 60M \\
    DW-T \cite{ren2022beyond} &  45.7   & -     & -     & 46.9  & 61M \\
    Focal-T \cite{yang2021focal} & 45.8 &  -    & -     & 47.0  & 62M \\
    {QFormer$_h$-T} & {46.9} & {58.3} & {82.1} &   {48.1}    & 61M \\
    \hline
    ResNet-101 \cite{he2016deep} & 43.8 & 54.7 & 81.0 & 44.9 & 86M\\
    Swin-S \cite{liu2021swin} & 47.6  & 58.8  & 82.5  & 49.5  & 81M \\
    DAT-S  \cite{xia2022vision} & 48.3  & 60.4  & -     & 49.8  & 81M \\
    Focal-S \cite{yang2021focal} & 48.0 & -     & -     & 50.0  & 85M \\
    {QFormer$_h$-S} & {48.9} & {60.3} & {83.0} &  {50.3}  & 82M \\
    \hline
    Swin-B \cite{liu2021swin} & 48.1 & 59.1 & - & 49.7 & 121M \\
    DW-B  \cite{ren2022beyond}  & 48.7 & -     & -     & 50.3  & 125M \\
    Focal-B \cite{yang2021focal} & 49.0 & - & - & 50.5 & 126M \\
    QFormer$_h$-B & 49.5 & 60.9 & 83.0 & 50.6 & 123M \\
    \hline
    \end{tabular}%
  \label{tab:ade20k}%
\end{table}%

\noindent\textbf{Results.} The results of QFormer$_h$-T with Mask RCNN and Cascade RCNN are presented in Table~\ref{tab:hierarchical MaskRCNN} and Table~\ref{tab:hierarchical Cascade RCNN}, respectively. More results of larger backbones, \eg, QFormer$_h$-S and QFormer$_h$-B, are presented in Table~\ref{tab:hierarchical Mask RCNN larger} and Table~\ref{tab:hierarchical Cascade RCNN larger}. As shown in Table~\ref{tab:hierarchical MaskRCNN}, QFormer$_h$-T with QA considerably outperforms the baseline method Swin-T~\cite{liu2021swin} in both object detection and instance segmentation tasks with the Mask RCNN detector. Specifically, QFormer$_h$-T improves the performance by 2.2 $mAP^{bb}$ and 1.7 $mAP^{mk}$ compared to Swin-T with the 1$\times$ training schedule. These findings confirm the efficacy of the proposed QA in handling objects of different scales which are very common in object detection datasets like MS COCO. Additionally, the performance of QFormer$_h$-T improves with an extended training schedule (3$\times$) and achieves a gain of 1.5 mAP$^{bb}$ and 1.1 mAP$^{mk}$ over Swin-T, showing that the proposed learnable QA continually benefits from a longer training schedule by fitting the data with diverse objects better. Furthermore, QA demonstrates superior performance compared to other advanced window attention methods, such as the multi-window approach (\ie, DW-T \cite{ren2022beyond}), obtaining a gain of 0.8 mAP$^{bb}$ with the 3$\times$ training schedule. This is because QA can learn quadrangles with arbitrary sizes and shapes to fit different objects, whereas the multiple windows used in DW-T are still manually designed, which requires careful tuning for each dataset and task. The superior performance of QA over the focal attention (\ie, Focal-T \cite{yang2021focal}) is also evident, with a performance gain of 1.1 mAP$^{bb}$ with the 1$\times$ training schedule. It is worth noting that stacking more transformer layers does not mitigate the inherent drawback of the hand-crafted designs of window attention, and the superiority of our QA remains. For instance, QFormer$_h$-S achieves the best performance among all other backbones, \ie, 49.5 mAP$^{bb}$ and 44.2 mAP$^{mk}$, respectively, as demonstrated in Table~\ref{tab:hierarchical Mask RCNN larger}.

The detection results of different backbones with the Cascade RCNN~\cite{cai2018cascade} detector are presented in Table~\ref{tab:hierarchical Cascade RCNN}. Under the 1$\times$ schedule, QFormer$_h$-T outperforms Swin-T with a significant margin of 1.7 mAP$^{bb}$ and $1.3$ mAP$^{mk}$. Moreover, as the training epochs increase, the proposed QA yields a performance gain of 1.2 mAP$^{bb}$ and 1.0 mAP$^{mk}$. For larger backbones such as QFormer$_h$-S and QFormer$_h$-B, we conduct a more detailed comparison regarding detection performance on objects of different scales, as shown in Table~\ref{tab:hierarchical Cascade RCNN larger}. The results demonstrate that QFormer$_h$-S and QFormer$h$-B exhibit substantially better performance compared to the Swin Transformers, with a minimum gain of 0.9 mAP$^{bb}$ and 0.7 mAP$^{mk}$. These findings demonstrate the potential of QA to enhance large models with stronger representation ability. Additionally, the results also reveal that QFormer$_h$ consistently outperforms Swin variants in terms of detection accuracy on objects of various scales. For example, the performance of Swin-B regarding small, middle, and large size objects, \ie, AP$^{bb}{s}$, AP$^{bb}{m}$, AP$^{bb}{l}$, is improved by 0.9 AP, 1.3 AP, and 1.1 AP, respectively, by employing the proposed QA. Such observations further validate the effectiveness of QA in handling objects of different sizes.

\subsubsection{Semantic segmentation}
\textbf{Settings.} In this section, we employ the ADE20k dataset \cite{zhou2017scene} to evaluate the performance of different backbones for the semantic segmentation task. Specifically, we adopt UPerNet~\cite{xiao2018unified} as the segmentation framework, following Swin Transformer \cite{liu2021swin}. All the models are trained and evaluated following common practice. We use the Adam optimizer with polynomial learning rate schedulers to train the models for 160k steps. The initial learning rate is initialized to 6e-5, and weight decay is set to 1e-2. All experiments are conducted on 8 NVIDIA A100 GPUs with a total batch size of 16. We provide both single-scale and multi-scale testing results.

\noindent\textbf{Results.} We report the results of QFormer$_h$ and other models in Table~\ref{tab:ade20k}. As can be seen, our QFormer$_h$ models perform the best when compared to all other backbones with hand-crafted window designs. For example, QFormer$_h$-T outperforms Swin-T by 2.4 mIoU, 2.7 mAcc, and 2.3 mIoU*, respectively. While Focal-T and DW-T allow for extracting long-range context and cross-window information exchange by attending to more tokens, their performance still falls behind the proposed QFormer$_h$-T, demonstrating the superiority of the learnable design of our QA over the hand-crafted windows. Furthermore, the benefit of QA extends to scaled-up backbones. For example, QFormer$_h$-S achieves 48.9 mIoU and 50.3 mIoU*, respectively, outperforming other backbones. It is also noteworthy that QFormer$_h$-S even outperforms the much larger Swin-B and DW-B models, showing that our QA can significantly improve the representation ability of vision transformers.

\begin{table}[htbp]
  \centering
  \caption{The inference speed of the backbones on different vision tasks. UPerNet \cite{zhou2017scene} and Cascade Mask RCNN \cite{cai2019cascade} are taken as the decoders for semantic segmentation and detection, respectively. The speed is measured on NVIDIA A100 GPUs. `WA' refers to window attention.}
    \begin{tabular}{c|c|c|c|c}
    \hline
    \multirow{2}[1]{*}{Backbone} & Speed  & \multirow{2}[1]{*}{Perform.} & \multirow{2}[1]{*}{Task} & \multirow{2}[1]{*}{Image size} \\
          & (fps) &       &       &  \\
    \hline
    Swin-T & 73.0  & 44.5 mIoU  & \multirow{10}[8]{*}{Segment} & \multirow{7}[6]{*}{(512,512)} \\
    QFormer$_h$-T & 63.4  & 46.9 mIoU &       &  \\
\cline{1-3}    Swin-S & 57.8  & 47.6 mIoU &       &  \\
    QFormer$_h$-S & 50.1  & 48.9 mIoU &       &  \\
\cline{1-3}    ViT-B + WA & 62.7  & 39.7 mIoU &       &  \\
    ViT-B + shifted WA & 62.1  & 41.6 mIoU &       &  \\
    QFormer$_p$-B & 58.9  & 43.6 mIoU &       &  \\
\cline{1-3}\cline{5-5}    ViT-B + WA & 45.1  & 40.2 mIoU  &       & \multirow{3}[2]{*}{(640,640)} \\
    ViT-B + shifted WA & 44.5  & 42.3 mIoU &       &  \\
    QFormer$_p$-B & 41.0  & 44.9 mIoU &       &  \\
    \hline
    Swin-T & 18.4  & 50.2 mAP & \multirow{4}[4]{*}{Detection} & \multirow{4}[4]{*}{(1344, 800)} \\
    QFormer$_h$-T & 16.1  & 51.4 mAP &       &  \\
\cline{1-3}    Swin-S & 15.7  & 51.9 mAP &       &  \\
    QFormer$_h$-S & 13.0  & 52.8 mAP &       &  \\
    \hline
    \end{tabular}%
  \label{tab:speed}%
\end{table}%

\subsection{Inference speed}

We evaluate the inference speed of plain and hierarchical architectures and present the results in Table~\ref{tab:speed}. In addition to the proposed QFormer, baseline models, namely Swin~\cite{liu2021swin} and ViT~\cite{vit}, have also been evaluated for comparison. For the tasks of semantic segmentation and object detection, we adopt UPerNet \cite{zhou2017scene} and Cascade Mask RCNN \cite{he2017mask} as task heads, respectively. The input image size is provided in the last column. We run each model 30 times and record the average speed. All experiments are conducted on NVIDIA A100 GPUs. As observed in the table, QFormer$_h$ is marginally slower than Swin Transformer by about 13\% for various vision tasks while achieving significantly better performance. For plain models, QFormer$_p$ reduces the speed gap to less than 9\% compared to ViT with window attention while obtaining a gain of over 4 mIoU, demonstrating the great potential of QA in achieving better Pareto frontier of speed and accuracy. The slightly slower speed of our model is attributed to the insufficient optimization of sampling operations compared with matrix multiplication operations in the PyTorch framework, where the latter is better optimized with cuBLAS. Faster speed can be achieved by integrating the sampling operation with subsequent attention and linear projection operations using CUDA acceleration techniques.

\subsection{Visualization and analysis}

\begin{figure}
    \centering
    \includegraphics[width=\linewidth]{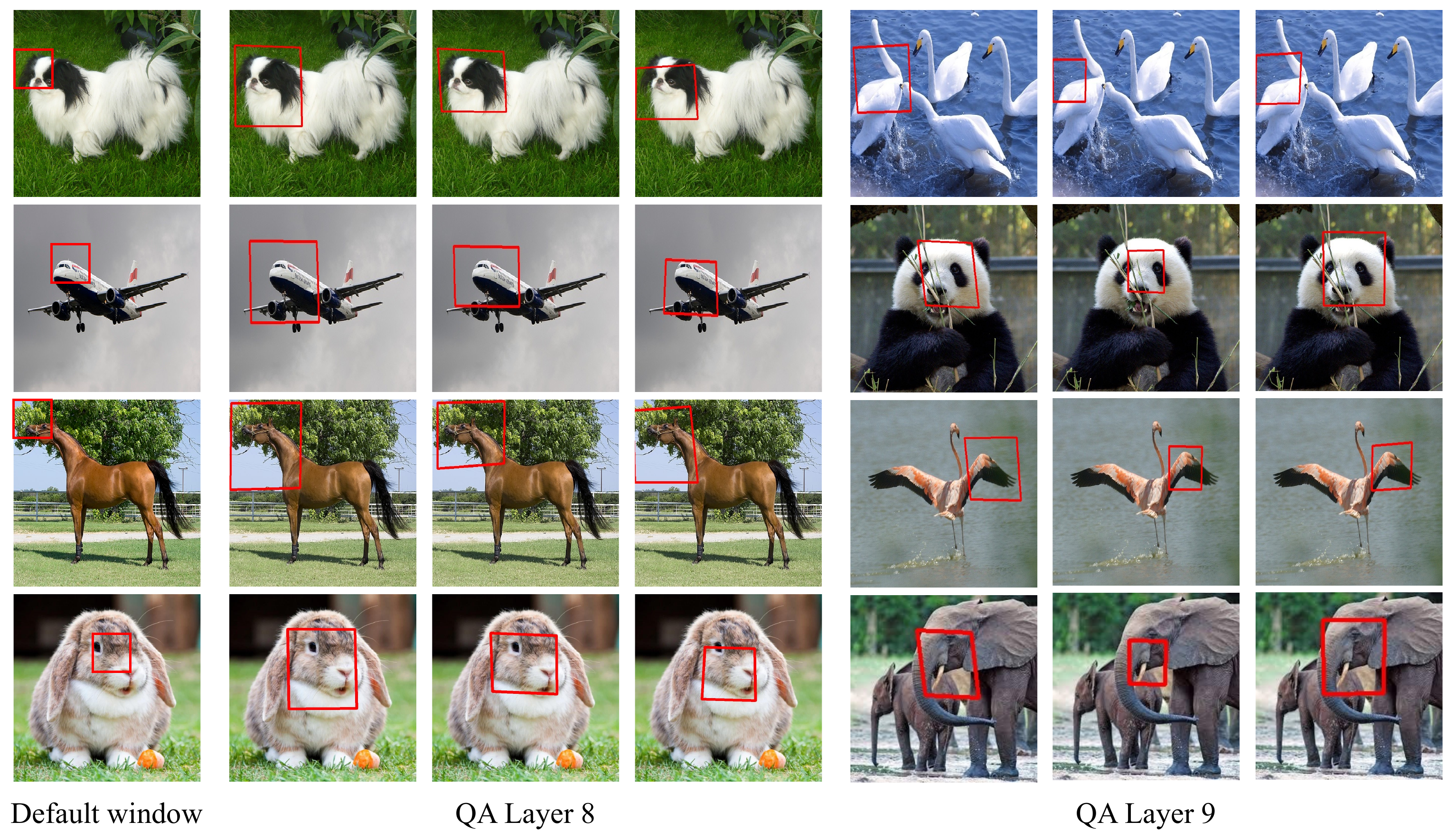}
    \caption{Visualization of the quadrangles generated by QFormer$_h$-T. The model is trained on ImageNet for classification.}
    \label{fig:cls window visualization}
\end{figure}

\begin{figure}
    \centering
    \includegraphics[width=\linewidth]{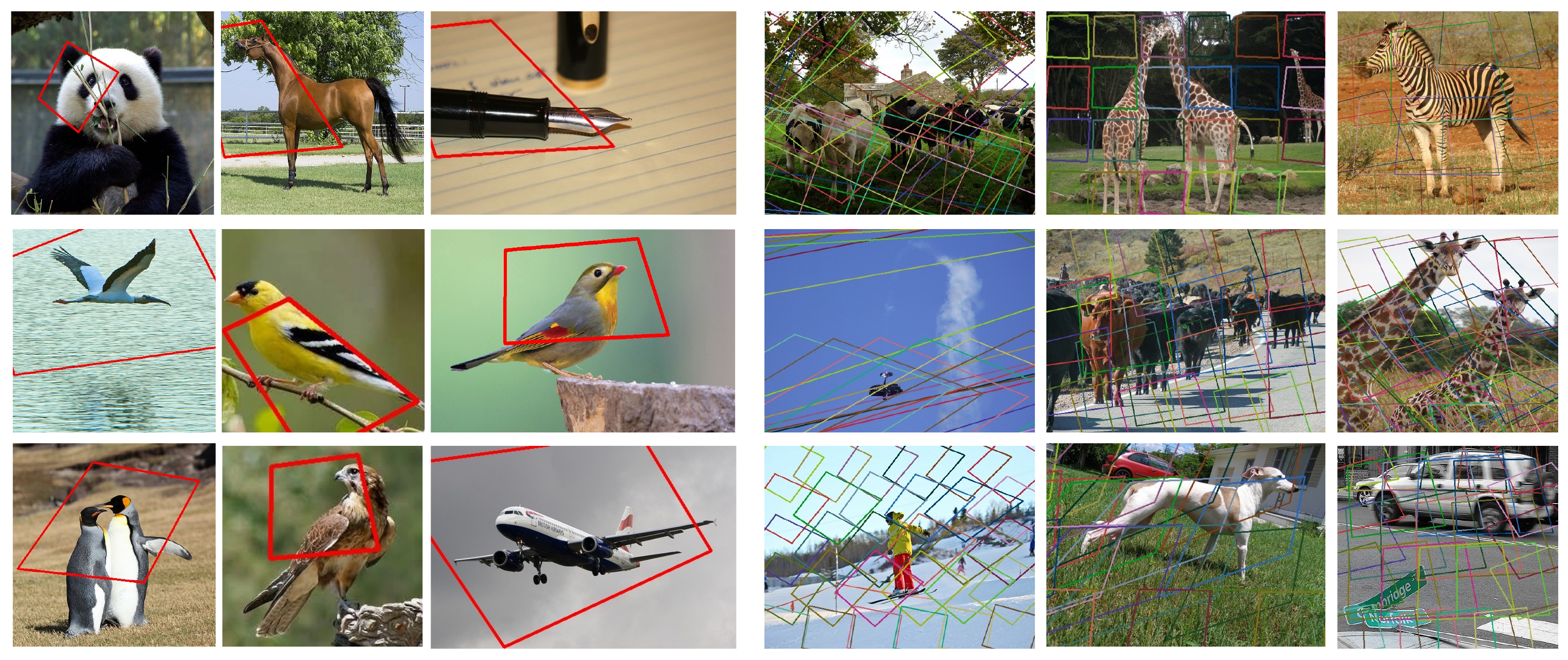}
    \caption{Visualization of the quadrangles generated by QFormer$_h$-T. The model is trained on MS COCO for object detection and instance segmentation.}
    \label{fig:det window visualization}
\end{figure}

\noindent\textbf{Visualization of learned quadrangles.} 
To examine how QA attends to various images, we visualize the default windows employed in Swin-T~\cite{liu2021swin} and the quadrangles generated by QA in Figure~\ref{fig:cls window visualization} and Figure~\ref{fig:det window visualization}. The images utilized are from the ImageNet~\cite{deng2009imagenet} and MS COCO~\cite{lin2014microsoft} datasets. The results presented in Figure~\ref{fig:cls window visualization} are derived from QFormer$_h$-T, which is trained on ImageNet for image classification, and the quadrangles in different attention heads are shown in distinct columns. The model used to obtain the results in Figure~\ref{fig:det window visualization} is trained on MS COCO.

The results in Figure~\ref{fig:cls window visualization} demonstrate that the quadrangles produced by QA can cover a more diverse set of regions within the target objects in the images compared to the default fixed-size windows, which can only cover a limited portion of the targets. For instance, the default window restricts the receptive field within the areas of dog and horse heads. In contrast, by learning the appropriate attention regions, QA expands the window size and captures a more comprehensive context of animal heads. Additionally, the quadrangles generated by different heads in QA exhibit varying sizes, locations, and shapes to attend to different parts of the targets, which facilitates capturing rich contextual information and improving object feature representations. Furthermore, the diverse quadrangles as illustrated in Figure~\ref{fig:det window visualization} promote cross-window information exchange, making it feasible to remove the shifted window operation that accompanies the window attention~\cite{liu2021swin}.

\begin{figure}
    \centering
    \includegraphics[width=\linewidth]{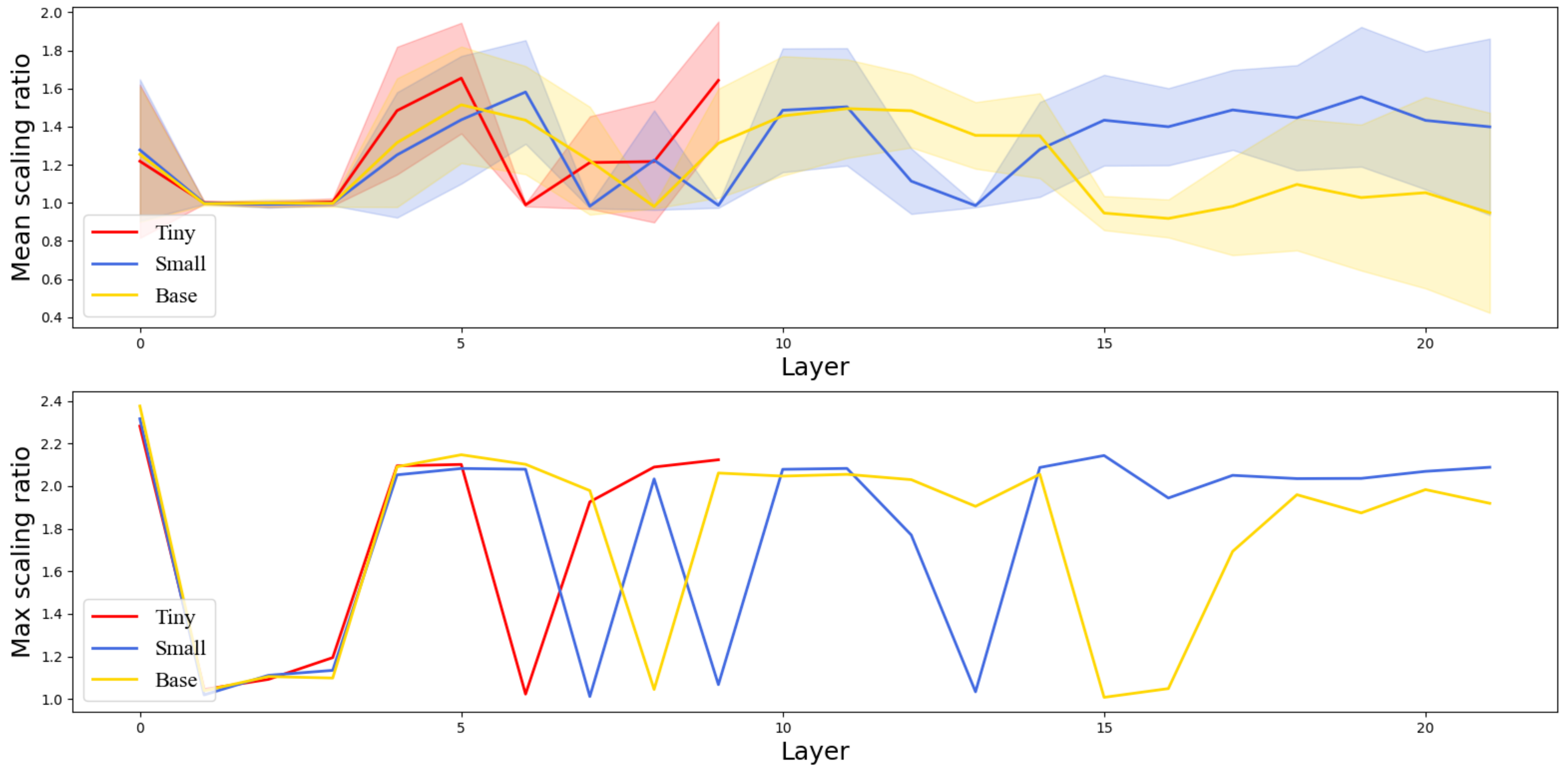}
    \caption{Visualization of the learned scale factors in QA. The mean (with standard deviation) and max values in each layer of all three hierarchical QFormer$_h$ models are presented.}
    \label{fig:scale}
\end{figure}

\noindent \textbf{Analysis of the learned transformations in QA.}
To explore the variations in attention regions across different layers, we perform an analysis of the learned parameters in the projective transformation $T$. To this end, we conduct experiments on the ImageNet validation set using three QFormer$_h$ models of varying depths and sizes. The results of these experiments are presented in Figure~\ref{fig:scale}, which illustrates the curves of the scaling factor learned by each layer. The upper figure shows the mean value with standard deviation, while the bottom figure shows the maximum value. Generally, the scaling factor plays a crucial role in determining the attention area. Notably, we observe that all other learned parameters, such as rotation, shearing, shift, and projection, have mean values that are close to zero, with moderate standard deviations.

Based on Figure~\ref{fig:scale}, it is evident that the first four layers of all three models exhibit a strong similarity in the learned scaling factors. Specifically, these layers have the nearly identical mean and maximum values. This suggests that the models tend to emphasize similar regions during the early stages of processing, regardless of their depth or width. Moreover, with the exception of the first layer, the mean scaling factors in the subsequent three layers are roughly equal to 1.0, representing the default window. This underscores the significance of local structures relative to global context during early-stage feature extraction, which echoes the earlier work on early convolution \cite{xiao2021early}. Conversely, in deeper layers, the attention area varies substantially across the different models, exhibiting significant layer-wise differences. Nevertheless, some commonalities can still be discerned. First, all three models exhibit layers with both large ($>1.0$) and small ($\sim$1.0) scaling factors, emphasizing the importance of utilizing diverse windows to gather information from both global and local contexts. Furthermore, by considering each layer with a small scaling factor as a starting point, i.e., layer 3 for all models and layers 6, 7, and 8 for QFormer$_h$-T, QFormer$_h$-S, and QFormer$_h$-B, respectively, the models appear to inherently comprise several stages. Within each stage, the mean and standard deviation of the scaling factors initially increase before decreasing.

\begin{figure}
    \centering
    \includegraphics[width=\linewidth]{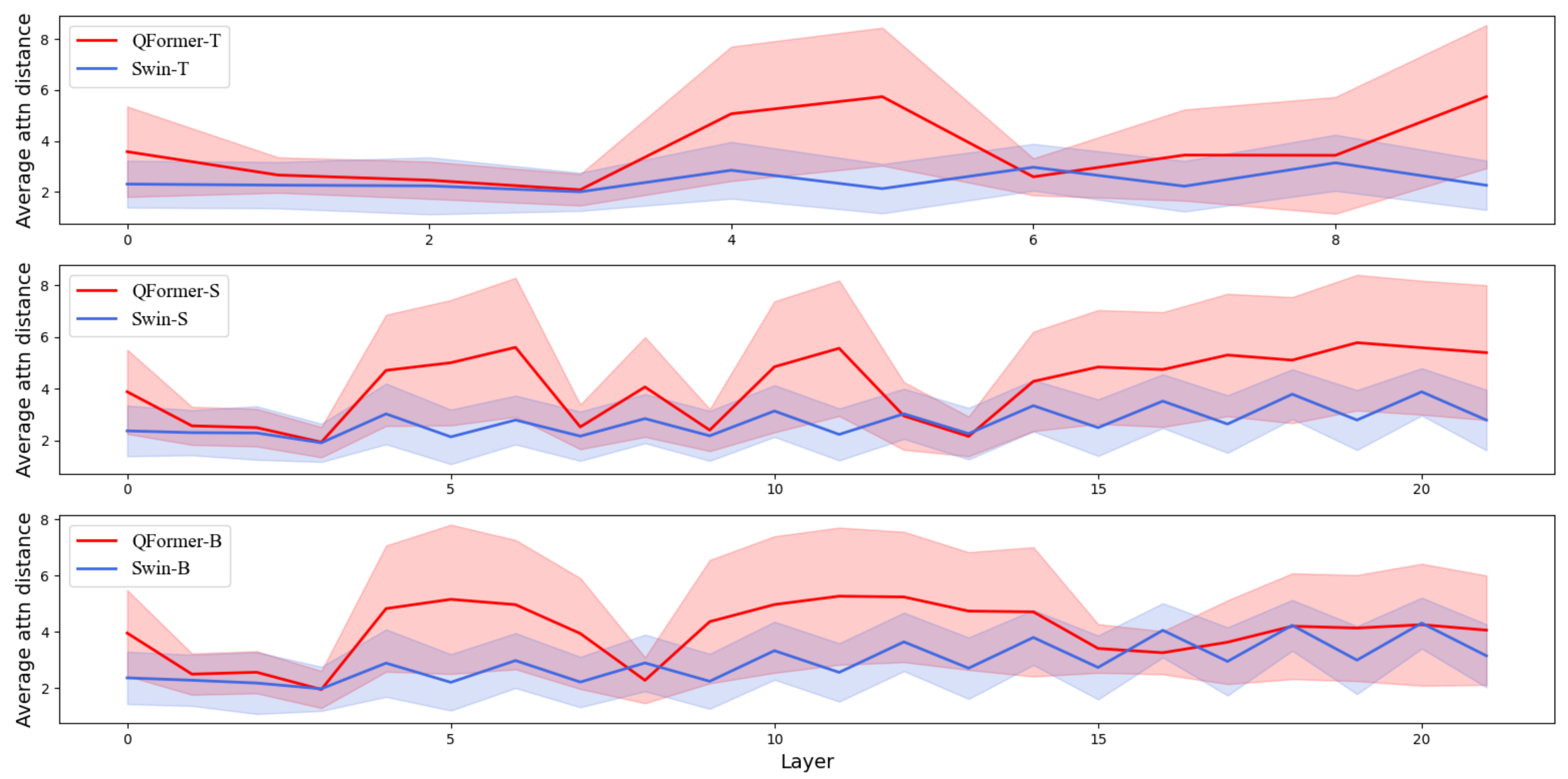}
    \caption{Visualization of attention distance. The mean (with standard deviation) attention distance of each layer in QFormer$_h$ and Swin Transformer is plotted.}
    \label{fig:attn distance}
\end{figure}

\noindent\textbf{Analysis of the real attention distance.} 
We compute the average attention distance (with standard deviation) for each layer of QFormer$_h$ and Swin Transformer. The attention distance is obtained by summing up the weighted pixel distance between any two query and key tokens, followed by averaging the score across all token pairs for each individual layer. The results are depicted in Figure~\ref{fig:attn distance}, with the curves for QFormer$_h$ and Swin Transformer denoted in red and blue, respectively. The figure reveals that our proposed QA exhibits significantly larger attention distances than the vanilla window attention in almost all layers for all three models. This finding supports our claim that QA is adept at adjusting the attention regions to handle objects of different sizes. Moreover, the larger standard deviation values in QA suggest a high degree of diversity in the attention regions, which can facilitate the capture of rich contextual information and the exchange of cross-window information. The stage-like variations of attention distance align with the observation in Figure~\ref{fig:scale}, while the attention distance of Swin Transformers remains largely consistent with marginal variations.

\section{Conclusion and Discussion}

In this paper, we introduce a novel quadrangle attention (QA) mechanism that learns attention regions from data, and integrate them into both plain and hierarchical vision transformers, resulting in a new architecture called QFormer. With minimal code modifications and negligible extra computation cost, QA provides the enhanced capacity to learn feature representation and effectively handle objects of varying sizes, shapes, and orientations, outperforming the vanilla window attention significantly. Our extensive experiments on various vision tasks, including image classification, object detection, instance segmentation, pose estimation, and semantic segmentation, demonstrate the efficacy of QA and the superiority of QFormer over representative models. Despite the promising results, there is still room for further improvement of QA and QFormer. For example, integrating QA into other training settings, such as self-supervised learning, warrants further research efforts. Moreover, in order to maintain the computational cost comparable to the vanilla window attention, we currently only sample sparse tokens from each target window, with the number of sampled tokens equaling the default window size. This may overlook some informative tokens as the window size increases, even though the missed information may be compensated for through feature exchange with other windows. Therefore, it is worth investigating other effective sampling strategies to improve model performance and inference speed.

\ifCLASSOPTIONcaptionsoff
  \newpage
\fi

\bibliographystyle{IEEEtran}
\bibliography{egbib}

\end{document}